\documentclass{article}

\usepackage[main,final]{neurips_2026}

\usepackage[utf8]{inputenc} 
\usepackage[T1]{fontenc}    
\usepackage{hyperref}       
\usepackage{url}            
\usepackage{booktabs}       
\usepackage{amsfonts}       
\usepackage{nicefrac}       
\usepackage{microtype}      
\usepackage{xcolor}         

\usepackage{orcidlink}
\usepackage{fontawesome5} 
\usepackage{adjustbox}

\usepackage{makecell}
\usepackage{comment}
\usepackage{amsmath}
\usepackage{multirow}
\usepackage[table,xcdraw]{xcolor}

\usepackage{wrapfig}
\usepackage[utf8]{inputenc} 
\usepackage[T1]{fontenc}    
\usepackage{nicefrac}       
\usepackage{bm}
\usepackage[dvipsnames]{xcolor}
\hypersetup{
  colorlinks,
  linkcolor={red!100!black},
  citecolor={blue!140!black},
  urlcolor={red!100!black}
  }
  \definecolor{orange}{HTML}{ff7f0e}
  \definecolor{blue}{HTML}{1f77b4}
\usepackage{enumitem}

\usepackage{subcaption}

\makeatletter
\@ifundefined{iflatexml}{\newif\iflatexml}{}
\makeatother
\title{PVRF: All-in-one Adverse Weather Removal via Prior-modulated and Velocity-constrained Rectified Flow}

\iflatexml
\author{%
Wei Dong\\
McMaster University
\And
Han Zhou\\
McMaster University\\
Corresponding Author, Project Lead
\And
Terry Ji\\
McMaster University
\And
Guanhua Zhao\\
McMaster University
\And
Shahab Asoodeh\\
McMaster University
\And
Yulun Zhang\\
Shanghai Jiao Tong University
\And
Guangtao Zhai\\
Shanghai Jiao Tong University
\And
Jun Chen\\
McMaster University
\And
Xiaohong Liu\\
Shanghai Jiao Tong University
}
\else
\author{Wei Dong$^{1}$, Han Zhou$^{\dagger1}$, Terry Ji$^{1}$, Guanhua Zhao$^{1}$, Shahab Asoodeh$^{1}$,\\ \textbf{Yulun Zhang$^{2}$,  Guangtao Zhai$^{2}$, Jun Chen$^{1}$, and Xiaohong Liu$^{2}$} \\
$^1$ McMaster University, $^2$ Shanghai Jiao Tong University \\
 $^\dagger$Corresponding Author, Project Lead
}
\fi

\begin{document}
\maketitle

{\centering
\newcommand{\capbox}[2]{%
  \parbox[t]{#1}{\centering\scriptsize #2}%
}

\newcommand{\percepbox}[2]{%
  \begingroup
  \setlength{\fboxsep}{0pt}%
  \noindent\colorbox{#1}{%
    \begin{minipage}[t]{\dimexpr\linewidth+5pt}
      #2
    \end{minipage}%
  }%
  \endgroup
}
\vspace{-8mm}

\begin{figure}[h]
\centering
    \includegraphics[width=\linewidth]{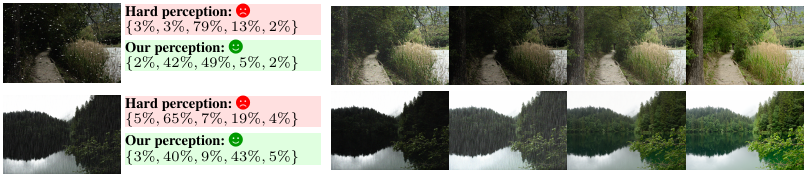}
    \noindent
\begin{minipage}[t]{0.3739\linewidth}
    \centering\scriptsize
    \capbox{\linewidth}{(i) Inputs + (hard vs.\ soft) priors}
\end{minipage}%
\hspace{0.02\linewidth}%
\begin{minipage}[t]{0.58\linewidth}
    \centering
    \begin{minipage}[t]{0.245\linewidth}
        \centering\scriptsize (ii) Backbone only
    \end{minipage}\hspace{0.002\linewidth}%
    \begin{minipage}[t]{0.245\linewidth}
        \centering\scriptsize (iii) w/ Hard prior
    \end{minipage}\hspace{0.002\linewidth}%
    \begin{minipage}[t]{0.245\linewidth}
        \centering\scriptsize (iv) w/ Ours prior
    \end{minipage}\hspace{0.002\linewidth}%
    \begin{minipage}[t]{0.245\linewidth}
        \centering\scriptsize (v) Ground Truth
    \end{minipage}
\end{minipage}
    \caption{\textbf{Mixed-degradation examples.} We compare restoration results with no prior (AdaIR~\cite{adair} as the backbone), hard one-hot conditioning, and our soft perception priors. Hard one-hot conditioning can be brittle under combined degradations, as it forces a single type decision. Soft priors are used as branch weights in the Weather-Weighted Adapter and lead to improved restoration under combined degradations. The ordering of the perception vector is: \{blur, low light, snow, rain, haze\}.}
    \label{fig:main_qual}

 \end{figure}
 }

\begin{abstract}

Adverse weather removal (AWR) in real-world images remains challenging due to heterogeneous and unseen degradations, while distortion-driven training often yields overly smooth results. We propose \textbf{PVRF}, a unified framework that integrates zero-shot soft weather \textbf{p}erceptions with \textbf{v}elocity-constrained \textbf{r}ectified-\textbf{f}low refinement. PVRF introduces an AWR-specific question answering module (AWR-QA) that uses frozen vision--language models (VLMs) to estimate soft probabilities of weather types and low-level attribute scores. These perceptions condition restoration networks via attribute-modulated normalization (AMN) and weather-weighted adapters (WWA), producing an anchor estimate for refinement. We then learn a terminal-consistent residual rectified flow with perception-adaptive source perturbation and a terminal-consistent velocity parameterization to stabilize learning near the terminal regime. Extensive experiments show that PVRF improves both fidelity and perceptual quality over state-of-the-art baselines, with strong cross-dataset generalization on single and combined degradations. Code will be released at \url{https://github.com/dongw22/PVRF}.

\end{abstract}
\section{Introduction}
\label{sec_intro}

As a fundamental task in computer vision, adverse weather removal (AWR) aims to restore weather-degraded images to clean counterparts, which is essential for  downstream visual perception, outdoor imaging, and robust scene understanding. Weather degradations include rain, snow, haze, and low-light conditions, among others~\cite{li2019heavy,glare, chen2021all,li2017aod,zhou2023breaking,dong2024dehazedct,dong2025sgllie,zhou2025lita,dong2026coscan}. While most prior works focus on a single degradation, recent studies have increasingly explored all-in-one blind AWR models that
adaptively handle diverse weather conditions within a single network~\cite{valanarasu2022transweather,ye2023adverse,yang2024language,PromptIR,luo2023controlling}. Representative backbones range from CNNs and Transformers to diffusion-based models~\cite{valanarasu2022transweather,adair,patch-diffusion,dong2024shadowrefiner}.

However, two challenges remain for real-world AWR and photo-realistic restoration. 
\textbf{First}, it is still non-trivial to consistently infer the underlying weather condition from degraded inputs and convert it into actionable conditioning signals. Existing approaches either encode weather cues implicitly via learnable architectures~\cite{PromptIR,valanarasu2022transweather,dong2024ecmamba} or predict discrete weather labels by training/fine-tuning classifiers and CLIP-based frameworks~\cite{UWADN,dcpt,jiang2024autodir,luo2023controlling,VLU-Net}. Such hard conditioning can be sensitive to cross-dataset appearance shifts and mixed/ambiguous degradations, which are common in practice (Fig.~\ref{fig:main_qual}). 
\textbf{Second}, distortion-centric discriminative networks often yields overly smooth outputs, whereas diffusion-based methods, although visually compelling, may underperform in fidelity. Achieving restorations that are both faithful and photo-realistic remains challenging in AWR.

To address these limitations, we propose \textbf{PVRF}, an all-in-one AWR framework guided by \emph{zero-shot soft weather perceptions} from pretrained vision--language models (VLMs). 
First, our AWR-specific question answering module (AWR-QA) uses concise, type-specific definitions and quantizes VLM responses into soft weather-type probabilities and low-level attribute scores. 
Second, we use these perceptions to modulate AWR backbones via Attribute-Modulated Normalization (AMN) and Weather-Weighted Adapters (WWA), enabling degradation-aware posterior estimation under distortion losses. 
More importantly, we introduce a perception-guided rectified flow that learns a transport from a perception-dependent, posterior-anchored source distribution toward the clean-image distribution. The proposed terminal-consistent residual formulation, together with perception-adaptive source perturbation, improves photo-realism while maintaining competitive fidelity (Fig.~\ref{fig_rf}).

In summary, PVRF contributes 
(i) definition-guided, zero-shot VLM perceptions yielding soft type and attribute priors, 
(ii) a perception-guided modulation design (AMN/WWA) for degradation-aware posterior estimation, and (iii) a terminal-consistent residual rectified flow with perception-adaptive source perturbation for photo-realistic refinement.

\begin{figure*}[t]
\setlength{\abovecaptionskip}{1mm}
\setlength{\parskip}{0mm} 
\setlength{\baselineskip}{0mm} 
\centering
\begin{minipage}[c]{1\textwidth}
    \centering
    \includegraphics[width = 0.86\textwidth]{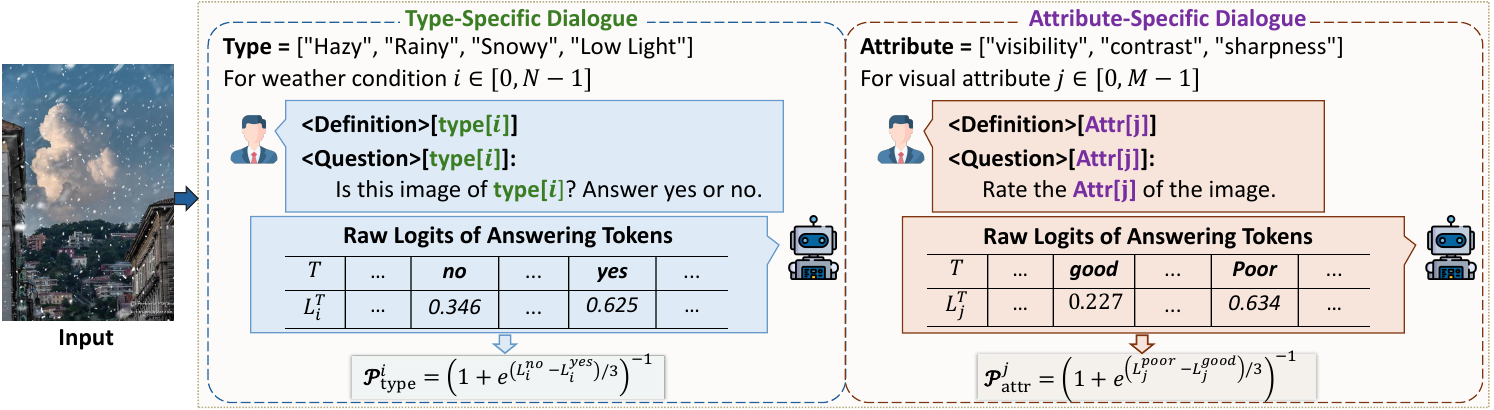}
\end{minipage}

\caption{Our developed AWR-specific Question Answering (AWR-QA) Module. We query the frozen VLM regarding each degradation or attribute independently to acquire robust degradation and low-level attribute perceptions. Other than hard/one-hot degradation indicators, our AWR-QA module supports inputs with combined degradations. }
\label{fig_QA}
\vspace{-6mm}
\end{figure*}

\section{Background}

\noindent{\textbf{Perception-Distortion Trade-off}}.
All-in-one adverse weather removal (AWR) aims to restore a clean image \(\bm{X}\sim p_{\bm{X}}\) from its degraded observation \(\bm{Y}\). The goal is to produce an estimate \(\hat{\bm{X}}\) that achieves both low distortion (\textit{e.g.}, high PSNR/SSIM) and high perceptual quality (\textit{i.e}., \(p_{\hat{\bm X}}\) matches \(p_{\bm X}\)). 
However, these objectives are fundamentally at odds due to the perception-distortion trade-off~\cite{blau2018perception}. The minimum MSE estimator \(\hat{\bm X}_{\text{mmse}}=\mathbb{E}[\bm X\mid \bm Y]\) minimizes distortion but yields over-smoothed results. In contrast, the ideal estimator \(\hat{\bm X}_{\text{opt}}\) is defined to minimize MSE subject to the perfect perceptual constraint \(p_{\hat{\bm X}}=p_{\bm X}\).

\noindent\textbf{Posterior-Mean Transport under Perceptual Constraints.} Freirich et al.~\cite{freirich2021theory} showed that \(\hat{\bm X}_{\text{opt}}\) can be characterized as transporting the posterior mean \(\hat{\bm X}_{\text{mmse}}\) toward the clean-image distribution \(p_{\bm X}\). Since rectified flow~\cite{liu2022flow} provides a practical means to learn optimal transport maps by training a velocity field \(v_{\bm \theta}\) on linear interpolation paths \(\bm Z_t = t\bm Z_1 + (1-t)\bm Z_0\) between coupled source and target samples \((\bm Z_0,\bm Z_1)\), 

PMRF~\cite{ohayon2024posterior} follows this formulation with \textit{\textbf{a two-stage pipeline}}: it first trains an MSE predictor \(f_{\bm \omega}\), and then fits a rectified-flow model from a noisy MMSE prediction to the clean target. Specifically, letting \(f_{\bm \omega^*}\) denote the trained posterior-mean predictor, PMRF defines:   

\begin{equation}
\bm{Z}_0 = f_{\bm \omega^*}(\bm{Y}) + \sigma_s \bm{\epsilon},\quad \bm{\epsilon} \sim \mathcal{N}(0, \bm{I}), \quad \bm{Z}_1 = \bm{X}, 
\label{eq:pmrf}
\end{equation}
where \(\sigma_s\) controls the perturbation magnitude.

\begin{figure*}[t]
\setlength{\abovecaptionskip}{1mm}
\setlength{\parskip}{0mm} 
\setlength{\baselineskip}{0mm} 
\centering
\begin{minipage}[c]{1\textwidth}
    \centering
    \includegraphics[width = 0.9\textwidth]{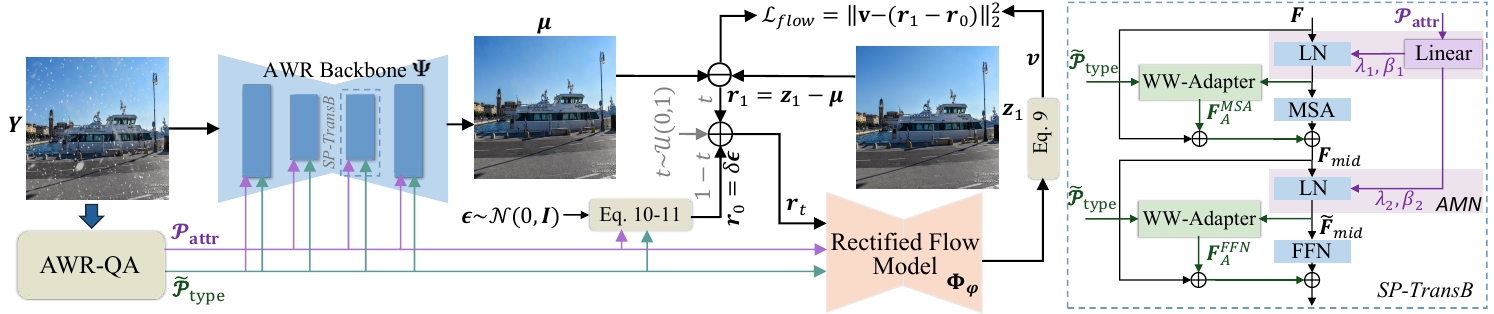}
\end{minipage}
\caption{The framework of \textbf{PVRF}. With soft perceptions ($\bm{\mathcal{P}}_{type}$ and $\bm{\mathcal{P}}_{attr}$) extracted from our developed AWR-QA, we introduce the Attribute-Modulated Normalization (AMN) and Weather-weighted Adapters (WWA) into AWR backbones to enhance fidelity. Subsequently, based on the posterior output $\bm \mu$, we develop a terminal-consistent residual flow model for photo-realistic restoration, where both the source distribution and velocity field estimation are informed by perceptions.}
\label{fig_framework}
\vspace{-4mm}
\end{figure*}

\section{Method}
\noindent{\textbf{Motivation}}.
Although Eq.~\ref{eq:pmrf} provides a posterior-mean transport approach for balancing distortion and perceptual quality, applying it directly to all-in-one adverse weather removal (AWR) is non-trivial due to the following challenges:
\begin{itemize}[label=\textbullet,leftmargin=*, labelsep=0.3em, itemsep=0.25em, topsep=0.25em]
\item \textbf{Mixed degradations yield a compromise posterior mean.} 
 
Training data in unified AWR spans multiple degradations $\mathcal{D}$.
Without any degradation cue, a single MSE predictor yields  
$\mathbb{E}[\bm X \mid \bm Y]
= \mathbb{E}_{\mathcal{\bm D}\sim p_{\text{train}}(\mathcal{\bm D}\mid \bm Y)}
\big[\,\mathbb{E}[\bm X\mid \bm Y,\mathcal{\bm D}]\,\big]$,

\textit{i.e.}, a weighted average of degradation-specific posterior means and thus a compromise estimate, which is exacerbated under distribution shift.
 
\item \textbf{Fixed perturbation is suboptimal.}
Gaussian perturbation is used to mitigate potential singularities in transport learning, but a fixed global scale $\sigma_s$ is ill-suited to all-in-one AWR with varying degradation severity. This fixed-scale design can \textit{over-perturb mild cases} and \textit{degrade fidelity}, while \textit{under-perturbing severe or ambiguous inputs} and \textit{providing insufficient regularization}. 

\item \textbf{Missing terminal constraint near $t\!\to\!1$.}
Rectified flow admits an implicit terminal relation: the optimal velocity field at $t=1$ satisfies
$\bm v^{\ast}(\bm X,1)=\bm X-\mathbb{E}[\bm Z_0\mid \bm Z_1=\bm X]$.
However, 
training over finitely sampled time points does not \emph{explicitly}

enforce this near $t\!\to\!1$. 

Crucially, under Eq.~\ref{eq:pmrf} the term $\mathbb{E}[\bm Z_0\mid \bm Z_1=\bm X]$ is generally intractable, making the terminal regime fragile.
\end{itemize}

To address these challenges, we first introduce AWR-specific Question Answering module to acquire reliable weather-type and low-level attribute perceptions from VLMs (Sec. \ref{QA}), then develop perception-guided conditioning designs for posterior estimation and velocity modeling (Sec. \ref{sec_2} and \ref{sec_RF}). Furthermore, in Sec. \ref{sec_RF}, we devise terminal-consistent residual flow with perception-adaptive source perturbation and terminal-consistent velocity parameterization. The framework of our method is presented in Fig. \ref{fig_framework}.

\subsection{Zero-shot Soft Perception Prior via VLM Question Answering}
\label{QA}

In unified AWR, the weather degradation is unknown and may be ambiguous or mixed.
A common way to introduce 
degradation 
cues is to predict a 
weather label and condition restoration on a one-hot indicator vector~\cite{luo2023controlling,dcpt,dong2026zero,dong2026towards}. This hard one-hot conditioning has three potential limitations: 
\textbf{training dependence} (new types often require retraining), 
\textbf{error sensitivity} (misclassification yields mismatched conditioning especially under cross-dataset shifts), 
and \textbf{a coarse cue} (discarding uncertainty and severity information).

To address these issues, we extract a \emph{zero-shot} and \emph{soft} perception prior from a frozen VLM via definition-augmented question answering (QA) (Fig.~\ref{fig_QA}), producing a type prior $\bm{\mathcal{P}}_{\text{type}}$ and an attribute prior $\bm{\mathcal{P}}_{\text{attr}}$.
For each weather type $i\in\{0,\dots,N-1\}$, we run an independent binary dialogue and quantify the preference between the canonical answer anchors \texttt{yes} and \texttt{no} using their raw logits:

\vspace{-5mm}
\begin{align}
\bm{\mathcal{P}}_{\text{type}} = \{\mathcal{P}^{i}_{\text{type}}\}, \quad \mathcal{P}^{i}_{\text{type}} = (1 + e^{(L_i^{\text{no}} - L_i^{\text{yes}}) / 3})^{-1}, \quad \bm{\tilde{\mathcal{P}}}_{\text{type}}=\frac{\bm{\mathcal{P}_{\text{type}}}}{\sum_{i=0}^{N-1}\mathcal{P}^{i}_{\text{type}}+\tau}, 
\label{eq:quantify_weather}
\end{align}
where $\tau = 10^{-8},\quad i \in [0, N-1].$ The resulting vector $\{\mathcal{P}^{i}_{\text{type}}\}$ retains uncertainty (multiple entries may be non-negligible), enabling weighted (rather than hard) conditioning downstream. We then normalize it onto the probability simplex $\bm{\tilde{\mathcal{P}}}_{\text{type}}$.

In addition to discrete types, we query $M$ low-level attributes (visibility, contrast, and sharpness) to provide severity cues that are often more stable under type ambiguity.
For each attribute $j\in\{0,\dots,M-1\}$, we anchor the answers with \texttt{good}/\texttt{poor} and compute:
\begin{align}
\bm{\mathcal{P}}_{\text{attr}} = \{\mathcal{P}^{j}_{\text{attr}}\}, \quad \mathcal{P}^{j}_{\text{attr}} = (1 + e^{(L_j^{\text{poor}} - L_j^{\text{good}}) / 3})^{-1}, \quad j \in [0, M-1]. 
\label{eq:quantify_attr}
\end{align}

Finally, to reduce confusion among visually overlapping degradations, we prepend a concise, type-specific \texttt{<Definition>[type[i]]} before the binary question, encouraging the VLM to judge causal semantics rather than only low-level artifacts.

\begin{figure*}[t]
\setlength{\abovecaptionskip}{1mm}
\setlength{\parskip}{0mm} 
\setlength{\baselineskip}{0mm} 
\centering
\begin{minipage}[c]{1\textwidth}
    \centering
    \includegraphics[width = 1\textwidth]{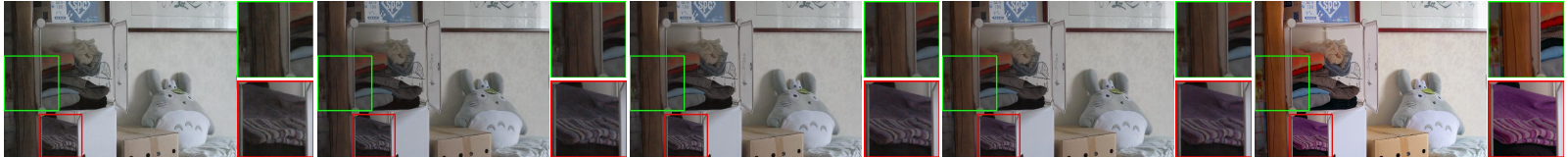}\\
    \begin{minipage}[c]{0.195\textwidth}
    \centering
    \scriptsize{Vanilla \\ PMRF}
    \end{minipage}
    \begin{minipage}[c]{0.195\textwidth}
    \centering
    \scriptsize{+ Adaptive \\ Perturbation}
    \end{minipage}
    \begin{minipage}[c]{0.195\textwidth}
    \centering
    \scriptsize{+ Perception-aware Velocity Modeling}
    \end{minipage}
    \begin{minipage}[c]{0.195\textwidth}
    \centering
    \scriptsize{+ TC Velocity \\ Parameterization}
    \end{minipage}
    \begin{minipage}[c]{0.195\textwidth}
    \centering
    \scriptsize{Ground \\ Truth}
    \end{minipage}
\end{minipage}
\caption{ Progressive Enhancement via Adaptive Perturbation, Perception-aware Velocity Modeling, and Terminal-Constraint (TC) Velocity Parameterization.}
\label{fig_rf}
\vspace{-4mm}
\end{figure*}

\subsection{Perception-Guided Conditioning Design}
\label{sec_2}
This section describes how the extracted soft perception priors are used to modulate network representations as conditioning signals.

A key consideration is that $\bm{\tilde{\mathcal{P}}}_{\text{type}}$ indicates \emph{which} weather degradations are present, while $\bm{\mathcal{P}}_{\text{attr}}$ provides continuous cues of \emph{how severe} the corruption is.
Accordingly, we incorporate them through two complementary mechanisms:
(i) attribute-modulated normalization for stable severity modulation, and
(ii) weather-weighted adapters for soft type-aware adaptation.

\noindent{\textbf{Attribute-Modulated Normalization (AMN)}.}

Since $\bm{\tilde{\mathcal{P}}}_{\text{attr}}$ reflects low-level severity and is less sensitive to type ambiguity, it is used to modulate normalization statistics via a feature-wise affine transformation.
Given an input feature $\mathbf{F}$ (and similarly an intermediate feature $\mathbf{F}_{\text{mid}}$), AMN predicts scale and bias from $\bm{\mathcal{P}}_{\text{attr}}$ and modulates the layer-normalized features:
\begin{align}
[\lambda_1, \beta_1, \lambda_2, \beta_2] = \text{Linear}(\bm{\mathcal{P}}_{\text{attr}}), \,
\tilde{\mathbf{F}} = \lambda_1 \odot \text{LN}(\mathbf{F}) + \beta_1, \,
 \tilde{\mathbf{F}}_{\text{mid}} = \lambda_2 \odot \text{LN}(\mathbf{F}_{\text{mid}}) + \beta_2,
\label{eq:AMN}
\end{align}
where $\odot$ denotes element-wise multiplication.
This provides a continuous, severity-aware modulation and mitigates error propagation when $\bm{\mathcal{P}}_{\text{type}}$ is uncertain.

\noindent
\begin{minipage}[t]{0.56\linewidth}
\vspace{0pt}
\noindent{\textbf{Weather-Weighted Adapter (WW-Adapter)}.}

To incorporate type information while preserving uncertainty, we use $\bm{\tilde{\mathcal{P}}}_{\text{type}}$ to softly weight multiple adapter 
branches. As shown in Fig.~\ref{fig:WW-Adapter}, the WW-Adapter contains $N$ branches, each corresponding to one weather type, and outputs: \vspace{-4mm}
\end{minipage}\hfill
\begin{minipage}[t]{0.42\linewidth}
\vspace{-5.0mm}
\centering
\includegraphics[width=\linewidth]{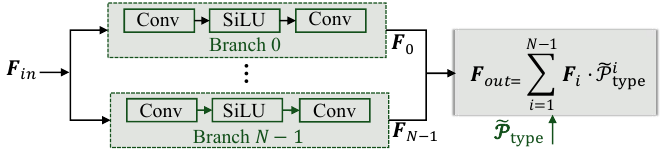}
\captionof{figure}{Soft weather perceptions $\tilde{\mathcal{P}}^{i}_{\text{type}}$ serve as branch weights in WW-Adapter. }
\label{fig:WW-Adapter}
\end{minipage}
\begin{align}
\vspace{-4mm}
\mathbf{F}_{\text{out}} = \sum_{i=0}^{N-1} \tilde{\mathcal{P}}^{i}_{\text{type}} \, \mathbf{F}_{i},
\qquad 
\mathbf{F}_{i} = [\text{Conv}, \text{SiLU}, \text{Conv}]^{i}(\mathbf{F}_{\text{in}}),
\label{eq:WW-Adapter}
\end{align}
where $\tilde{\mathcal{P}}^{i}_{\text{type}}$ serves as the soft mixture weight for the $i$-th branch. 
Compared with a hard single-type conditioning, this weighted aggregation naturally supports mixed or ambiguous conditions and reduces sensitivity to misclassification.

\begin{table*}[t]
  \centering
  \scriptsize
  \caption{Comparison with SOTA methods on all-in-one image restoration (setting I: Haze + Rain + Snow). The \textcolor{red}{Best}, \textcolor{blue}{Second-best}, and \textcolor{orange}{Third-best} methods are highlighted.\vspace{-2mm}}
  \label{tab_setting1}
  \centering
  \setlength\tabcolsep{1pt}
  \renewcommand{\arraystretch}{1.1}
  \resizebox{\textwidth}{!}{%
  \begin{tabular}{c|c|cc|ccccccc|ccc|c}
    \toprule[1pt]
    \multirow{2}{*}{Task} & \multirow{2}{*}{Metrics} & \multicolumn{2}{c|}{CNN-based} & \multicolumn{7}{c|}{Transformer-based}  & \multicolumn{3}{c|}{SDE/Diffusion-based} & \textbf{Our} \\
 
    & & WGWS-Net & DCPT-NAFNet & SwinIR & PromptIR & TransWeather & Histoformer & GridFormer & AdaIR & HOGFormer & DACLIP &GPPLLIE & UniRestore  &\textbf{PVRF} \\
    \midrule
    \multirow{8}{*}{\textit{Dehazing}}
    & PSNR$\uparrow$     &25.2422 &29.5123 &28.1335 &29.0742 &28.8708 &28.8753 &\textcolor{orange}{29.7015} &\textcolor{blue}{29.8552} &26.2312 &29.1248 &28.9203 &24.1892 & \textcolor{red}{31.7364} \\
    & SSIM$\uparrow$     & 0.9198 & \textcolor{blue}{0.9561} & 0.9490 & 0.9492 & 0.9449 & 0.9532 & \textcolor{orange}{0.9550} & 0.9523 & 0.9446 & 0.9366 & 0.9498 & 0.8272 & \textcolor{red}{0.9592} \\
    & LPIPS$\downarrow$  & 0.0915 & \textcolor{blue}{0.0245} & 0.0351 & 0.0302 & 0.0327 & 0.0284 & \textcolor{orange}{0.0249} & 0.0290 & 0.0361 & \textcolor{orange}{0.0249} & 0.0336 & 0.1008 & \textcolor{red}{0.0217} \\
    & FID$\downarrow$    &13.9170 & 6.6464 &10.5504 & 8.5657 & 9.6171 & 8.1998 & \textcolor{blue}{5.4800} & 8.2878 & 9.7512 & \textcolor{orange}{5.6563} & 9.3160 &19.1667 & \textcolor{red}{4.1366} \\
    & MUSIQ$\uparrow$    &52.3578 &56.1662 &55.9391 &55.9365 &56.0669 &56.1384 &\textcolor{orange}{56.2397} &55.9506 &56.0219 &\textcolor{blue}{56.871}  &55.8657 &54.7502 &\textcolor{red}{57.4256} \\
    & CLIPIQA$\uparrow$  & 0.2506 & 0.2580 & 0.2944 & \textcolor{blue}{0.3046} & 0.2831 & 0.2591 & 0.2616 & 0.2957 & 0.2557 & 0.2623 & \textcolor{orange}{0.3045} & 0.2530 & \textcolor{red}{0.3102} \\
    & NIQE$\downarrow$   & 4.3388 & 4.0909 & 3.9766 & 3.9844 & \textcolor{orange}{3.8393} & 4.0926 & 3.9909 & 4.0457 & 4.1116 & \textcolor{blue}{3.8933} & 3.9641 & 4.4329 & \textcolor{red}{3.5843} \\
    & MANIQA$\uparrow$   & 0.6277 & 0.6398 & 0.6390 & \textcolor{orange}{0.6420} & 0.6367 & 0.6379 & 0.6416 & \textcolor{blue}{0.6423} & 0.6359 & 0.6360 & 0.6408 & 0.6087 & \textcolor{red}{0.6511} \\

    \midrule
    \multirow{8}{*}{\textit{Deraining}}
    & PSNR$\uparrow$     &25.4435 &26.6555 &28.1922 &\textcolor{orange}{28.2102} &24.3043 &27.2624 &26.9545 &\textcolor{blue}{28.2166} &25.9042 &26.8019 &27.3112 &21.8189 &\textcolor{red}{28.4622} \\
    & SSIM$\uparrow$     & 0.8057 & 0.8340 & \textcolor{orange}{0.8870} &\textcolor{blue}{0.8881} & 0.8151 & 0.8551 & 0.8397 & 0.8868 & 0.8226 & 0.8504 & 0.8784 & 0.6732 & \textcolor{red}{0.8887} \\
    & LPIPS$\downarrow$  & 0.1922 & 0.1536 & 0.0748 & \textcolor{orange}{0.0739} & 0.1365 & 0.1331 & 0.1066 & \textcolor{blue}{0.0738} & 0.1626 & 0.0776 & 0.0967 & 0.2254 & \textcolor{red}{0.0542} \\
    & FID$\downarrow$    &42.6285 &54.8526 &\textcolor{blue}{25.3864} &25.9583 &50.3034 &44.4640 &34.9925 &\textcolor{orange}{25.4323} &57.4706 &29.3110 &30.4662 &67.9509 &\textcolor{red}{20.8527} \\
    & MUSIQ$\uparrow$    &63.9255 &64.9781 &70.4030 &\textcolor{blue}{70.7116} &68.8380 &67.7278 &68.7697 &\textcolor{orange}{70.5658} &66.0232 &70.3025 &70.2249 &64.6621 &\textcolor{red}{71.2542} \\
    & CLIPIQA$\uparrow$  & 0.6303 & 0.6326 & 0.7695 & \textcolor{orange}{0.7827} & 0.6721 & 0.6856 & 0.7025 & \textcolor{blue}{0.7871} & 0.6623 & 0.7589 & 0.7359 & 0.6585 & \textcolor{red}{0.7995} \\
    & NIQE$\downarrow$   & 3.5515 & 3.5908 & 3.6381 & 3.6483 & 3.7658 & 3.4542 & 3.4429 & 3.5867 & \textcolor{orange}{3.3721} & \textcolor{red}{3.1197} & 3.5928 & 4.5751 & \textcolor{blue}{3.1712} \\
    & MANIQA$\uparrow$   & 0.6208 & 0.6119 & \textcolor{orange}{0.6859} & \textcolor{blue}{0.6871} & 0.6253 & 0.6417 & 0.6383 & \textcolor{blue}{0.6871} & 0.6149 & 0.6827 & 0.6798 & 0.6135 & \textcolor{red}{0.7024} \\

    \midrule
    \multirow{8}{*}{\textit{Desnowing}}
    & PSNR$\uparrow$     &26.2841 &27.5454 &27.4389 &\textcolor{orange}{28.2108} &26.9457 &27.3890 &27.5816 &\textcolor{blue}{28.4096} &26.2566 &27.0387 &27.6699 &21.8825 &\textcolor{red}{28.6066} \\
    & SSIM$\uparrow$     & 0.8330 & 0.8886 & 0.8874 & \textcolor{orange}{0.8966} & 0.8769 & 0.8870 & 0.8890 & \textcolor{blue}{0.8992} & 0.8794 & 0.8701 & 0.8938 & 0.6402 & \textcolor{red}{0.9005} \\
    & LPIPS$\downarrow$  & 0.1145 & 0.0808 & 0.0705 & \textcolor{orange}{0.0627} & 0.0790 & 0.0846 & 0.0734 & \textcolor{blue}{0.0590} & 0.0886 & 0.0677 & 0.0668 & 0.1145 & \textcolor{red}{0.0526} \\
    & FID$\downarrow$    &27.2584 &29.1007 &25.1864 &\textcolor{orange}{22.1090} &29.7221 &30.7080 &23.6043 &\textcolor{blue}{20.8006} &32.8779 &23.4294 &25.3291 &40.0907 &\textcolor{red}{16.6937} \\
    & MUSIQ$\uparrow$    &69.0165 &69.7187 &68.6227 &69.1966 &69.1384 &\textcolor{orange}{70.1404} &69.1025 &69.2825 &69.0672 &\textcolor{red}{70.6822} &69.4237 &68.5674 &\textcolor{blue}{70.6515} \\
    & CLIPIQA$\uparrow$  & 0.4787 & 0.4836 & 0.5552 & \textcolor{orange}{0.5638} & 0.5362 & 0.4947 & 0.5017 & \textcolor{blue}{0.5663} & 0.4864 & 0.4819 & 0.5452 & 0.4484 & \textcolor{red}{0.5912} \\
    & NIQE$\downarrow$   & 3.1464 & 3.0592 & \textcolor{blue}{2.8550} & 2.9555 & 2.9998 & 3.1293 & \textcolor{orange}{2.8889} & 2.9380 & 3.0156 & 2.9284 & 3.0459 & 3.3874 & \textcolor{red}{2.6396} \\
    & MANIQA$\uparrow$   & 0.6645 & 0.6611 & 0.6688 & 0.6749 & 0.6584 & 0.6651 & 0.6681 & \textcolor{orange}{0.6771} & 0.6601 & \textcolor{blue}{0.6829} & 0.6698 & 0.6356 & \textcolor{red}{0.6949} \\

    \midrule
    \multirow{8}{*}{\textit{Average}}
    & PSNR$\uparrow$    & 25.6122 & 28.5387 & 27.9082 & \textcolor{orange}{28.6901} & 27.7213 & 28.2002 & 28.6890 & \textcolor{blue}{29.1908} & 26.2034 & 28.1707 & 28.3243 & 23.1562 & \textcolor{red}{30.3284} \\
    & SSIM$\uparrow$    & 0.8782  & 0.9200  & 0.9216  & \textcolor{orange}{0.9249}  & 0.9078  & 0.9202  & 0.9202  & \textcolor{blue}{0.9273}  & 0.9093  & 0.9048  & 0.9232  & 0.7477  & \textcolor{red}{0.9318} \\
    & LPIPS$\downarrow$ & 0.1104  & 0.0576  & 0.0513  & 0.0459  & 0.0597  & 0.0588  & 0.0502  & \textcolor{blue}{0.0440}  & 0.0677  & \textcolor{orange}{0.0450}  & 0.0517  & 0.1192  & \textcolor{red}{0.0356} \\
    & FID$\downarrow$   & 21.5575 & 19.4927 & 17.0820 & 15.0166 & 20.8444 & 19.7380 & 14.8055 & \textcolor{orange}{14.3673} & 22.7679 & \textcolor{blue}{14.2141} & 17.0083 & 31.5666 & \textcolor{red}{10.1833} \\
    & MUSIQ$\uparrow$   & 59.2015 & 61.6673 & 61.7779 & 62.0022 & 61.8471 & \textcolor{orange}{62.0979} & 61.9235 & 62.0225 & 61.4858 & \textcolor{blue}{62.9714} & 61.9846 & 60.4618 & \textcolor{red}{63.3748} \\
    & CLIPIQA$\uparrow$ & 0.3689  & 0.3749  & 0.4342  & \textcolor{blue}{0.4442}  & 0.4108  & 0.3851  & 0.3907  & \textcolor{orange}{0.4406}  & 0.3778  & 0.3907  & 0.4327  & 0.3632  & \textcolor{red}{0.4583} \\
    & NIQE$\downarrow$  & 3.8535  & 3.6911  & 3.5647  & 3.6037  & \textcolor{orange}{3.5510}  & 3.7002  & 3.5623  & 3.6251  & 3.6637  & \textcolor{blue}{3.4854}  & 3.6165  & 4.0998  & \textcolor{red}{3.2232} \\
    & MANIQA$\uparrow$  & 0.6392  & 0.6438  & 0.6542  & \textcolor{orange}{0.6580}  & 0.6427  & 0.6474  & 0.6501  & \textcolor{blue}{0.6589}  & 0.6416  & 0.6568  & 0.6548  & 0.6182  & \textcolor{red}{0.6714} \\

  \bottomrule[1pt]
  \end{tabular}
  }
  \vspace{-5mm}
\end{table*}
\subsection{Terminal-Consistent Residual Flow}
\label{sec_RF}

Given a degraded input $\mathbf{Y}$ and its soft prior $\bm{c}=\{\bm{\tilde{\mathcal{P}}}_{\text{type}},\bm{\mathcal{P}}_{\text{attr}}\}$, we first obtain a posterior-mean anchor $\bm{\mu}=\mathbf{\Psi^*}(\mathbf{Y},c)$, where $\mathbf{\Psi^*}(\cdot)$ denotes the trained perception-conditioned AWR backbone. 
We define the source sample as $\bm Z_0=\bm{\mu}+\delta\,\bm{\epsilon}$ with $\bm{\epsilon}\sim\mathcal{N}(0,\bm{I})$ and $\delta$ controlling the perturbation scale. In the PMRF setting (Eq.~\ref{eq:pmrf}), the terminal relation involves $\mathbb{E}[\bm Z_0\mid \bm Z_1=\bm X]$, where $\bm X$ denotes the clean target image. This term requires averaging the predictor output over the unknown conditional distribution $p(\bm Y\mid \bm X)$ and is therefore generally intractable to compute or enforce explicitly in AWR.

To obtain a more tractable formulation and improve stability near $t\!\to\!1$, we reparameterize the transport in a posterior-anchored residual space by defining $\bm r_0=\bm Z_0-\bm{\mu}$ and $\bm r_1=\bm Z_1-\bm{\mu}$, so that $\mathbb{E}[\bm r_0\mid \bm r_1]=0$ holds by construction. This residual formulation removes the content-dependent shift from the terminal relation and enables a terminal-consistent velocity parameterization.
The linear interpolation path in residual space is then:
\begin{equation}
\bm r_t = (1-t)\bm r_0 + t \bm r_1, \quad t \in [0,1].
\end{equation}

\noindent{\textbf{Terminal-Consistent Velocity Parameterization.}} Instead of directly regressing the residual displacement $\bm r_1 - \bm r_0$, 
we parameterize the residual-space velocity using a correction network $\mathbf{\Phi_{\bm{\varphi}}}(\bm r_t, t; \bm{\tilde{\mathcal{P}}}_{\text{type}}, \bm{\mathcal{P}}_{\text{attr}})$, which predicts a correction field in Eq.~\ref{eq:terminal_consistent_velocity}. Since the coefficient $t(1-t)$ vanishes at $t=1$, the velocity reduces to $v(r_1, 1) = r_1$ regardless of $\mathbf{\Phi}_{\bm{\varphi}}$. Thus, the terminal behavior is guaranteed by the parameterization, rather than being implicitly fitted by the network.
\begin{equation}
\bm v_{\bm{\varphi}}(\bm r_t, t;\bm{\tilde{\mathcal{P}}}_{\text{type}}, \bm{\mathcal{P}}_{\text{attr}}) = t\bm r_t + t(1-t) \, \mathbf{\Phi}_{\bm{\varphi}}(\bm r_t, t; \bm{\tilde{\mathcal{P}}}_{\text{type}}, \bm{\mathcal{P}}_{\text{attr}}).
\label{eq:terminal_consistent_velocity}
\end{equation}

\noindent{\textbf{Perception-adaptive Source Perturbation.}} In unified AWR, a single global perturbation scale is often mismatched to the input difficulty: mild cases can be over-perturbed (hurting fidelity), while severe or ambiguous cases may be under-regularized (making transport learning unstable). We therefore let the soft perceptions determine the perturbation scale of the rectified-flow source distribution. We derive two scalar descriptors from the perceptions: a normalized type-uncertainty score $H$ and an attribute-based severity score $s_{\text{attr}}$:
\begin{align}
H &= -\frac{\sum_{i=0}^{N-1}\tilde{\mathcal{P}}^i_{\text{type}}\log \tilde{\mathcal{P}}^i_{\text{type}}}{\log N},
\qquad
s_{\text{attr}} = 1 - \frac{1}{M}\sum_{j=0}^{M-1}\mathcal{P}^j_{\text{attr}},
\label{eq:prior_descriptors}
\end{align}
These descriptors are then fused into a scalar difficulty score $u\in[0,1]$, which is mapped to a perception-adaptive perturbation scale:
\begin{align}
u &= \alpha H + (1-\alpha)s_{\text{attr}},
\qquad
\delta = \delta_{\min} + (\delta_{\max}-\delta_{\min})u,
\label{eq:entropy}
\end{align}
where $\alpha$ is set to 0.5 to control the trade-off between weather uncertainty and attribute severity. We bound $\delta$ within $[\delta_{\min},\delta_{\max}]$ to balance data-consistency around the anchor and sufficient stochasticity for stable transport learning. In all experiments, we set $\delta_{\min}=0.025$ and $\delta_{\max}=0.1$.

\noindent{\textbf{Training Objective.}}
Following rectified flow, the target velocity is $\bm r_1-\bm r_0$. 
We train the correction network $\mathbf{\Phi}_{\bm{\varphi}}$ through the terminal-consistent parameterization in Eq.~\ref{eq:terminal_consistent_velocity} by minimizing:

\begin{equation}
\mathcal{L}_{\text{flow}} = \mathbb{E} \left[ \left\| \bm{v}_{\bm{\varphi}}(\bm r_t, t; \bm{\tilde{\mathcal{P}}}_{\text{type}}, \bm{\mathcal{P}}_{\text{attr}}) - (\bm r_1 - \bm r_0) \right\|_2^2 \right],
\label{eq:flow_loss}
\end{equation}
where $t$ is sampled from a uniform distribution $\mathcal{U}(0,1)$.

\begin{figure*}[t]
\setlength{\abovecaptionskip}{1mm}
\setlength{\parskip}{0mm} 
\setlength{\baselineskip}{0mm} 
\centering
\begin{minipage}[c]{1\textwidth}
    \centering
    \includegraphics[width = 1\textwidth]{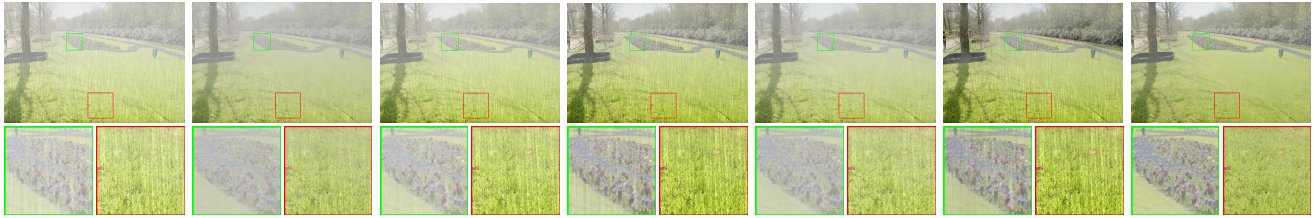}
    \includegraphics[width = 1\textwidth]{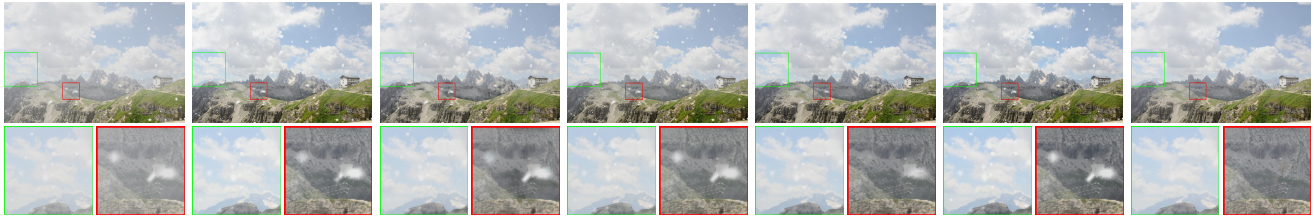}
    \begin{minipage}[c]{0.12\textwidth}
    \vspace{-1mm}
    \centering
    \scriptsize{Input}
    \end{minipage}
    \begin{minipage}[c]{0.14\textwidth}
    \vspace{-1mm}
    \centering
    \scriptsize{PromptIR}
    \end{minipage}
    \begin{minipage}[c]{0.14\textwidth}
    \vspace{-1mm}
    \centering
    \scriptsize{Histoformer}
    \end{minipage}
    \begin{minipage}[c]{0.14\textwidth}
    \vspace{-1mm}
    \centering
   \scriptsize{HOGformer}
    \end{minipage}
    \begin{minipage}[c]{0.14\textwidth}
    \vspace{-1mm}
    \centering
    \scriptsize{AdaIR}
    \end{minipage}
    \begin{minipage}[c]{0.14\textwidth}
    \vspace{-1mm}
    \centering
    \scriptsize{BioIR}
    \end{minipage}
    \begin{minipage}[c]{0.12\textwidth}
    \vspace{-1mm}
    \centering
   \scriptsize{\textbf{Ours}}
    \end{minipage}
\end{minipage}
\caption{Visual comparisons on inputs with \textit{combined degradations}. Baseline methods typically address one degradation. In contrast, our method handles multiple degradations simultaneously and produces outputs with enhanced clarity. Top: Haze + Rain; Bottom: Haze + Snow.}
\label{fig_setting3_combined}
\vspace{-2mm}
\end{figure*}


\section{Experiments}
\label{sec_exp}

\begin{table*}[t]
  \centering
  \scriptsize
  \caption{Quantitative comparisons on 5-task general image restoration (setting III). \vspace{-2mm}}
  \label{tab_setting3}
  \centering
  \setlength\tabcolsep{2pt}
  \renewcommand{\arraystretch}{1.0}
  \resizebox{0.82\textwidth}{!}{%
  \begin{tabular}{c|c|ccccc|c}
    \toprule[1pt]
    \multirow{2}{*}{Tasks} & \multirow{2}{*}{Metrics} & \multicolumn{5}{c|}{Baselines} &  \textbf{Ours} \\

    & & PromptIR\cite{PromptIR}  & Histoformer\cite{histoformer} & BioIR\cite{BioIR} & AdaIR\cite{adair} & HOGFormer\cite{hogformer} &\textbf{PVRF} \\
    
    \midrule
    \multirow{5}{*}{Deblurring}
    & PSNR$\uparrow$  &\textcolor{blue}{27.7319} &26.2834 &27.3538 &27.5489 &\textcolor{orange}{27.6222}  &\textcolor{red}{28.8893}   \\
    & SSIM$\uparrow$   &\textcolor{blue}{0.8441} &0.8066  &0.8322  &\textcolor{orange}{0.8384}  &0.8379   &\textcolor{red}{0.8606} \\
    & LPIPS$\downarrow$&0.1477 &0.2880  &0.1474  &\textcolor{orange}{0.1460}  &\textcolor{blue}{0.1441}   &\textcolor{red}{0.1006} \\
    & MUSIQ$\uparrow$  &36.8191 &28.4345 &\textcolor{blue}{39.0496} &\textcolor{orange}{37.1750} &37.0950  &\textcolor{red}{39.7532} \\
    & NIQE$\downarrow$ &5.0368 &5.9943  &\textcolor{orange}{4.5947}  &4.8374  &\textcolor{blue}{4.4739}   &\textcolor{red}{4.1958}\\
    \midrule
    \multirow{5}{*}{Low-light}
    & PSNR$\uparrow$      &21.5092  &21.4153  &\textcolor{orange}{22.0378}  &\textcolor{blue}{22.2035}  &21.3092  &\textcolor{red}{23.3552}   \\
    & SSIM$\uparrow$      &\textcolor{orange}{0.8297}   &0.8156   &0.8223   &\textcolor{blue}{0.8349}   &0.8143   &\textcolor{red}{0.8716} \\
    & LPIPS$\downarrow$   &\textcolor{orange}{0.1071}   &0.1496   &0.1214   &\textcolor{blue}{0.1016}   &0.1533   &\textcolor{red}{0.0917}   \\
    & MUSIQ$\uparrow$     &\textcolor{orange}{71.2539}  &65.8724  &70.3303  &\textcolor{blue}{71.6560}  &68.5276  &\textcolor{red}{72.4661}  \\
    & NIQE$\downarrow$    &4.2755   &\textcolor{blue}{4.1152}   &4.3271   &4.4665   &\textcolor{red}{4.0377}   &\textcolor{orange}{4.1956}    \\

    \midrule
    \multirow{5}{*}{Desnowing}
    & PSNR$\uparrow$      &\textcolor{orange}{27.5732}  &26.8895  &27.3654  &\textcolor{blue}{27.6996}  &27.0249  &\textcolor{red}{28.9248}  \\
    & SSIM$\uparrow$      &\textcolor{blue}{0.8886}   &0.8848   &0.8868   &\textcolor{orange}{0.8881}   &0.8857   &\textcolor{red}{0.8942} \\
    & LPIPS$\downarrow$   &\textcolor{blue}{0.0694}   &0.0805   &0.0717   &\textcolor{orange}{0.0700}   &0.0784   &\textcolor{red}{0.0515} \\
    & MUSIQ$\uparrow$     &69.0062  &\textcolor{blue}{69.4893}  &68.9797  &\textcolor{orange}{69.1223}  &68.7923  &\textcolor{red}{70.2035} \\
    & NIQE$\downarrow$    &\textcolor{blue}{2.9151}   &3.0640   &2.9237   &\textcolor{red}{2.9150}   &2.9949   &\textcolor{orange}{2.9169} \\

    \midrule
    \multirow{5}{*}{Deraining}
    & PSNR$\uparrow$      &\textcolor{orange}{31.0913}   &30.1356   &30.6917   &\textcolor{blue}{31.2414}   &30.8576   &\textcolor{red}{32.0514}   \\
    & SSIM$\uparrow$      &\textcolor{orange}{0.8944}    &0.8782    &0.8894    &\textcolor{blue}{0.8957}    &0.8913    &\textcolor{red}{0.9029} \\
    & LPIPS$\downarrow$   &\textcolor{orange}{0.0621}    &0.1006    &0.0684    &\textcolor{blue}{0.0616}    &0.0994    &\textcolor{red}{0.0538} \\
    & MUSIQ$\uparrow$     &\textcolor{orange}{65.8239}   &64.5085   &65.6481   &\textcolor{blue}{65.8400}   &64.8286   &\textcolor{red}{66.7914} \\
    & NIQE$\downarrow$    &3.5121    &3.5732    &\textcolor{blue}{3.4484}    &\textcolor{orange}{3.4837}    &3.4850    &\textcolor{red}{3.3468} \\

    \midrule
    \multirow{5}{*}{Dehazing}
    & PSNR$\uparrow$     &\textcolor{orange}{28.1200}  &27.3440  &27.0922  &\textcolor{blue}{28.2959}  &27.0393   &\textcolor{red}{29.5607}   \\
    & SSIM$\uparrow$     &\textcolor{blue}{0.9438}   &0.9415   &0.9409   &\textcolor{orange}{0.9436}   &0.9388    &\textcolor{red}{0.9516} \\
    & LPIPS$\downarrow$  &\textcolor{blue}{0.0332}   &0.0381   &0.0358   &\textcolor{orange}{0.0338}   &0.0389    &\textcolor{red}{0.0285} \\
    & MUSIQ$\uparrow$    &\textcolor{orange}{55.9653}  &55.1809  &\textcolor{blue}{56.0036}  &55.8192  &55.3643   &\textcolor{red}{56.5730} \\
    & NIQE$\downarrow$   &3.9656   &4.0448   &\textcolor{red}{3.9138}   &4.0052   &\textcolor{blue}{3.9234}    &\textcolor{orange}{3.9557} \\
  \bottomrule[1pt]
  \end{tabular}
  }
  \vspace{-3mm}
\end{table*}

\subsection{Experiment Setup and Implementation Details}
We evaluate PVRF under three all-in-one experimental settings. Setting I (3-task adverse weather removal) involves three core weather degradation tasks: dehazing, deraining, and desnowing. Setting II further introduces low-light enhancement as an additional domain. In Setting I, all approaches are trained on a unified dataset comprising Reside-6K \cite{reside6k}, Rain100H \cite{rain100L}, and Snow100K-L \cite{snow100k}, while LOLv2-Real \cite{LOLv2} is included in Setting II. Evaluations for the first two settings are conducted on the corresponding test splits. Setting III focuses on 5-task general image restoration, including deblurring (GoPro \cite{Gopro}), low-light enhancement (LOL \cite{LOLv1}), desnowing (Snow100K-L \cite{snow100k}), deraining (Merged rain dataset \cite{mergerain}), and dehazing (Reside-6K \cite{reside6k}). In addition to evaluating the test data corresponding to the training samples, we test on unseen datasets (HIDE \cite{hide}, MEF \cite{MEF}, NPE \cite{NPE}, DICM \cite{DICM}, RealRain-1k \cite{realrain}, and CDD-11 \cite{cdd_dataset}) and images with combined degradations (CDD-11 \cite{cdd_dataset}) to evaluate generalization.

\noindent \textbf{Baselines and Metrics.} We benchmark our method against state-of-the-art (SOTA) approaches for all-in-one image restoration tasks, including CNN-based methods (NAFNet \cite{nafnet}, WGWS-Net \cite{wswg}, DCPT-NAFNet \cite{dcpt}), Transformer-based methods (SwinIR \cite{liang2021swinir}, PromptIR \cite{PromptIR}, TransWeather \cite{valanarasu2022transweather}, Histoformer \cite{histoformer}, GridFormer \cite{gridformer}, AdaIR \cite{adair}, BioIR \cite{BioIR}, HOGformer \cite{hogformer}), and SDE/Diffusion-based methods (DACLIP \cite{luo2023controlling}, GPPLLIE \cite{gppllie}, and UniRestore \cite{chen2025unirestore}). For evaluation, we report fidelity metrics (PSNR, SSIM \cite{wang2004image}) and perceptual metrics (LPIPS \cite{zhang2018unreasonable}, FID \cite{heusel2017gans}, MUSIQ \cite{ke2021musiq}, CLIPIQA \cite{wang2023exploring}, NIQE \cite{mittal2012making}, MANIQA \cite{yang2022maniqa}).

\noindent \textbf{Implementation Details.} There are two stages in our implementation: (1) training our developed VLM-modulated AWR model; (2) optimizing $\mathbf{\Phi}_{\bm{\varphi}}$ using Eq.~\ref{eq:flow_loss}. For both stages, the total number of epochs is set to 200. The initial learning rate is set to $2 \times 10^{-4}$ and is reduced to $6.25 \times 10^{-6}$ by the end of training. Horizontal flips and rotations are used for data augmentation, and the batch size is set to 4. Each training input is cropped to $192 \times 192$ for Settings I and II, and to $128 \times 128$ for Setting III. For fair comparisons, we re-train all baselines for 400 epochs using one H100 while keeping other configurations (\textit{i.e.}, batch size, crop size, and learning rate schedule) unchanged.

\begin{figure*}[t]
\setlength{\abovecaptionskip}{1mm}
\setlength{\parskip}{0mm} 
\setlength{\baselineskip}{0mm} 
\centering
\begin{minipage}[c]{1\textwidth}
    \centering
    \includegraphics[width = 1\textwidth]{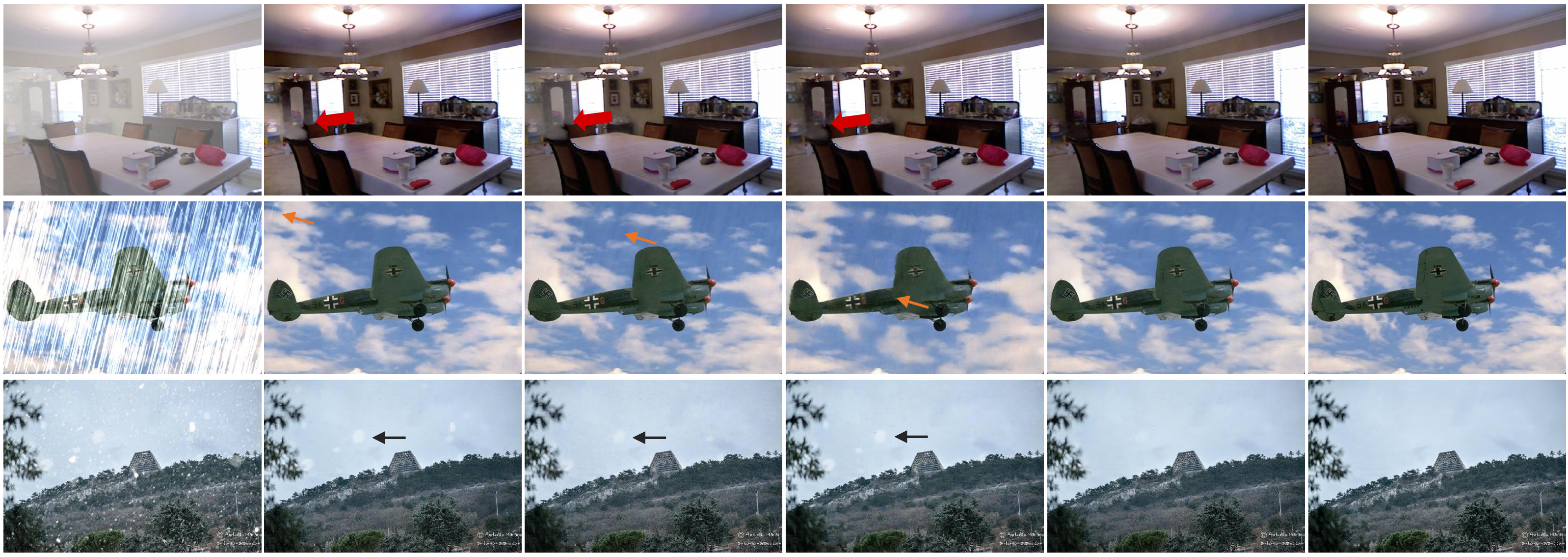}
    \begin{minipage}[c]{0.15\textwidth}
    \vspace{-1mm}
    \centering
    \scriptsize{Input}
    \end{minipage}
    \begin{minipage}[c]{0.15\textwidth}
    \vspace{-1mm}
    \centering
    \scriptsize{SwinIR}
    \end{minipage}
    \begin{minipage}[c]{0.15\textwidth}
    \vspace{-1mm}
    \centering
    \scriptsize{PromptIR}
    \end{minipage}
    \begin{minipage}[c]{0.15\textwidth}
    \vspace{-1mm}
    \centering
    \scriptsize{DCPT}
    \end{minipage}
    \begin{minipage}[c]{0.15\textwidth}
    \vspace{-1mm}
    \centering
    \scriptsize{\textbf{PVRF}}
    \end{minipage}
    \begin{minipage}[c]{0.15\textwidth}
    \vspace{-1mm}
    \centering
    \scriptsize{GT}
    \end{minipage}
    
\end{minipage}

\caption{Visual comparisons on setting I. Our PVRF is capable of preserving fine details and maintaining perceptual quality. need revision}
\label{fig_setting1}
\vspace{-3mm}
\end{figure*}

\begin{figure*}[t]
\setlength{\abovecaptionskip}{1mm}
\setlength{\parskip}{0mm} 
\setlength{\baselineskip}{0mm} 
\centering
\begin{minipage}[c]{1\textwidth}
    \centering
    \includegraphics[width = 1\textwidth]{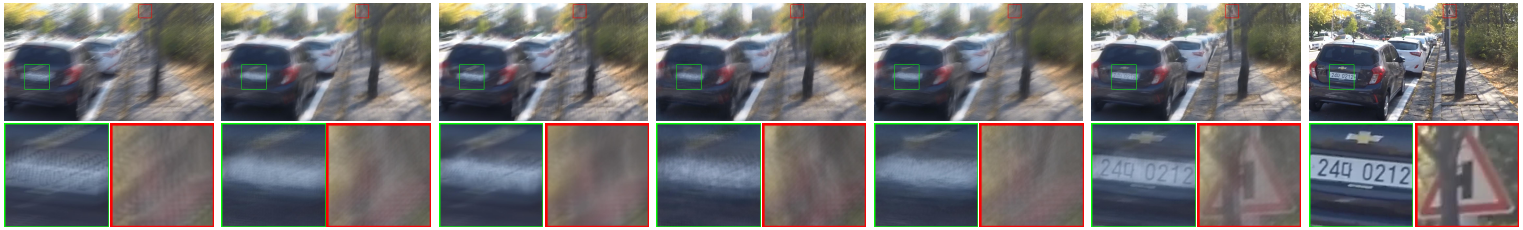}
    \includegraphics[width = 1\textwidth]{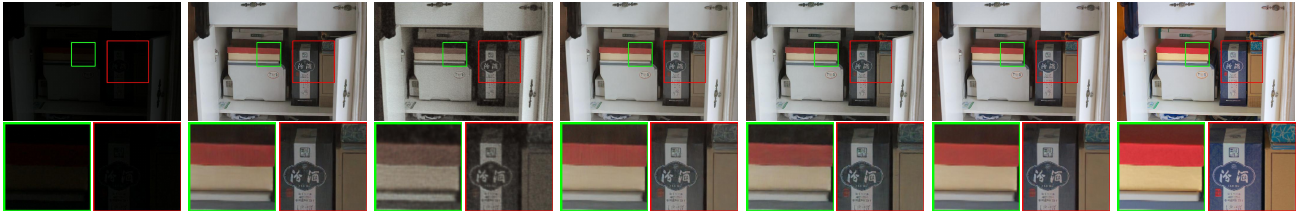}
    \includegraphics[width = 1\textwidth]{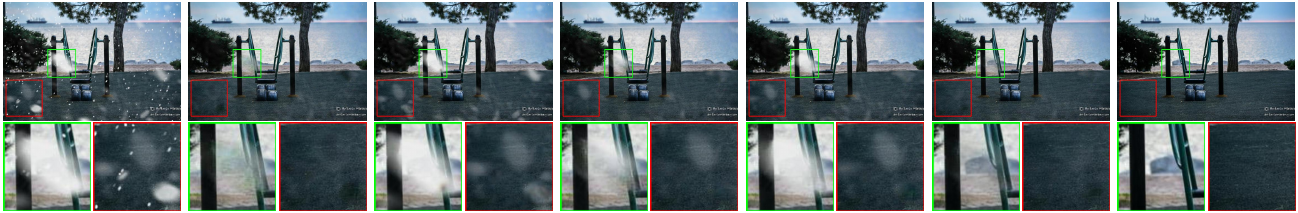}
    \begin{minipage}[c]{0.13\textwidth}
    \vspace{-1mm}
    \centering
    \scriptsize{Input}
    \end{minipage}
    \begin{minipage}[c]{0.13\textwidth}
    \vspace{-1mm}
    \centering
    \scriptsize{PromptIR}
    \end{minipage}
    \begin{minipage}[c]{0.139\textwidth}
    \vspace{-1mm}
    \centering
    \scriptsize{Histoformer}
    \end{minipage}
    \begin{minipage}[c]{0.139\textwidth}
    \vspace{-1mm}
    \centering
    \scriptsize{AdaIR}
    \end{minipage}
    \begin{minipage}[c]{0.139\textwidth}
    \vspace{-1mm}
    \centering
    \scriptsize{BioIR
    }
    \end{minipage}
    \begin{minipage}[c]{0.13\textwidth}
    \vspace{-1mm}
    \centering
    \scriptsize{\textbf{Ours}}
    \end{minipage}
    \begin{minipage}[c]{0.13\textwidth}
    \vspace{-1mm}
    \centering
    \scriptsize{GT}
    \end{minipage}
\end{minipage}
\caption{Visual comparisons on Setting III. Compared to other methods, our outputs are more visually appealing and deliver higher fidelity.}
\label{fig_setting3}
\vspace{-4mm}
\end{figure*}

\subsection{Comparisons with Baselines}
We report comparisons for Settings I and III in this section. Refer Appendix for Setting II results.

\noindent \textbf{Quantitative Results.} Tab.~\ref{tab_setting1} summarizes the quantitative comparisons on Setting I. Our method achieves superior performance on all three tasks, highlighting its advantage. Notably, the PSNR of our method surpasses the best SOTA by 1.14 dB on average. Moreover, our perceptual metrics significantly outperform the CNN-based and Transformer-based baselines, with only a few metrics slightly below DACLIP, a generative approach that requires 100 inference steps. Tab.~\ref{tab_setting3} reports quantitative comparisons on Setting III, where PVRF also achieves the best performance on both fidelity and perceptual metrics. These fidelity results demonstrate the effectiveness of the soft perceptions extracted from VLMs and our designed AMN and WWA modules, while the exceptional perceptual quality corroborates the importance of our terminal-consistent residual flow model.

\begin{table*}[!t]
  \centering
  \scriptsize
  \caption{Generalization performance on unseen data with single degradation. All methods are pre-trained on setting III (5-task general image restoration). \vspace{-1mm}}
  \label{tab_setting3_unseen}
  \centering
  \setlength\tabcolsep{3pt}
  \renewcommand{\arraystretch}{1.1}
  \resizebox{0.83\textwidth}{!}{%
  \begin{tabular}{c|c|ccccc|c}
    \toprule[1pt]
    \multirow{2}{*}{Tasks} & \multirow{2}{*}{Metrics} & \multicolumn{5}{c|}{Baselines} &  \textbf{Ours} \\

    & & PromptIR\cite{PromptIR}  & Histoformer\cite{histoformer} & BioIR\cite{BioIR} & AdaIR\cite{adair} & HOGFormer\cite{hogformer} &\textbf{PVRF} \\
    
    \midrule
    \multirow{3}{*}{\makecell{Deblurring\\HIDE~\cite{hide}}}
    & MUSIQ$\uparrow$    &54.3664  &32.9367 &\textcolor{orange}{54.4542} &\textcolor{blue}{54.8978}  &40.4104  &\textcolor{red}{55.1979} \\
    & NIQE$\downarrow$   &4.2547   &5.5078  &\textcolor{blue}{4.0421}  &\textcolor{orange}{4.0695}   &4.8670   &\textcolor{red}{3.7618} \\
    & MANIQA$\uparrow$   &\textcolor{orange}{0.2270}   &0.1180  &\textcolor{blue}{0.2459}  &0.2230   &0.1525   &\textcolor{red}{0.2468} \\
    
    \midrule
    \multirow{3}{*}{\makecell{Low-light\\MEF~\cite{MEF}, NPE~\cite{NPE} \\DICM~\cite{DICM}}}
    & MUSIQ$\uparrow$    &\textcolor{orange}{62.5308}  &58.3451  &\textcolor{blue}{62.6905}  &61.7768  &62.0512  &\textcolor{red}{62.9187} \\
    & NIQE$\downarrow$   &3.6653   &3.7302   &\textcolor{orange}{3.4984}   &3.6716   &\textcolor{blue}{3.4914}   &\textcolor{red}{3.4777} \\
    & MANIQA$\uparrow$   &\textcolor{blue}{0.4117}   &0.3491   &0.4030   &\textcolor{orange}{0.4052}   &0.4045   &\textcolor{red}{0.4140} \\

    \midrule
    \multirow{3}{*}{\makecell{Desnowing\\CDD-11~\cite{cdd_dataset}}}
    & MUSIQ$\uparrow$    &\textcolor{orange}{70.3515}  &69.5630  &69.8984   &\textcolor{blue}{70.5599}  &68.6738  &\textcolor{red}{70.7852} \\
    & NIQE$\downarrow$   &4.8778   &\textcolor{red}{4.5605}   &\textcolor{orange}{4.6191}    &4.6473   &4.6631   &\textcolor{blue}{4.5967} \\
    & MANIQA$\uparrow$   &\textcolor{orange}{0.5564}   &0.5263   &0.5502    &\textcolor{blue}{0.5573}   &0.5446   &\textcolor{red}{0.5703} \\

    \midrule
    \multirow{3}{*}{\makecell{Deraining\\RealRain-1k~\cite{realrain}}}
    & MUSIQ$\uparrow$    &\textcolor{blue}{39.5539}  &38.6795  &39.4703 &38.7431  &\textcolor{orange}{39.5328}  &\textcolor{red}{40.7962} \\
    & NIQE$\downarrow$   &6.8331   &7.6892   &\textcolor{orange}{6.5092}  &6.6111   &\textcolor{blue}{6.4646}   &\textcolor{red}{6.1126} \\
    & MANIQA$\uparrow$   &0.2106   &\textcolor{blue}{0.2222}   &0.2084  &0.2061   &\textcolor{orange}{0.2116}   &\textcolor{red}{0.2377} \\

    \midrule
    \multirow{3}{*}{\makecell{Dehazing\\CDD-11~\cite{cdd_dataset}}}
    & MUSIQ$\uparrow$    &70.5079  &69.3672   &\textcolor{orange}{70.6928}   &\textcolor{blue}{70.6985}  &70.6220  &\textcolor{red}{71.1582} \\
    & NIQE$\downarrow$   &\textcolor{red}{4.4472}   &4.5255    &\textcolor{blue}{4.4642}    &4.8394   &4.7481   &\textcolor{orange}{4.5051} \\
    & MANIQA$\uparrow$   &0.5703   &0.4772    &\textcolor{orange}{0.5704}    &\textcolor{blue}{0.5831}   &0.5684   &\textcolor{red}{0.5843} \\
  \bottomrule[1pt]
  \end{tabular}
  }
  \vspace{-2mm}
\end{table*}

\begin{table*}[t]
  \centering
  \scriptsize
  \caption{Generalization performance on images with combined degradations. \vspace{-1mm}}
  \label{tab_setting3_combined}
  \centering
  \setlength\tabcolsep{3pt}
  \renewcommand{\arraystretch}{1}
  \resizebox{0.82\textwidth}{!}{%
  \begin{tabular}{c|c|ccccc|c}
    \toprule[1pt]
    \multirow{2}{*}{Tasks} & \multirow{2}{*}{Metrics} & \multicolumn{5}{c|}{Baselines} &  \textbf{Ours} \\

    & & PromptIR\cite{PromptIR}  & Histoformer\cite{histoformer} & BioIR\cite{BioIR} & AdaIR\cite{adair} & HOGFormer\cite{hogformer} &\textbf{PVRF} \\
    
    \midrule
    \multirow{3}{*}{Haze + Rain \cite{cdd_dataset}}
    & MUSIQ$\uparrow$    &68.8037  &68.2682  &\textcolor{orange}{69.4325}  &69.3709  &\textcolor{blue}{69.9626}  &\textcolor{red}{70.5356} \\
    & NIQE$\downarrow$   &4.5900   &\textcolor{orange}{4.3552}   &4.4414   &\textcolor{blue}{4.3248}   &4.8599   &\textcolor{red}{4.289} \\
    & MANIQA$\uparrow$   &\textcolor{orange}{0.5055}   &0.4662   &\textcolor{blue}{0.5063}   &0.5035   &0.4945   &\textcolor{red}{0.5377} \\
    
    \midrule
    \multirow{3}{*}{Haze + Snow  \cite{cdd_dataset}}
    & MUSIQ$\uparrow$    &67.6534  &65.0106  &\textcolor{orange}{67.7196}  &\textcolor{blue}{67.7704}  &66.7434  &\textcolor{red}{68.9478} \\
    & NIQE$\downarrow$   &4.4024   &\textcolor{orange}{4.3374}   &\textcolor{red}{4.2164}   &4.5142   &4.5555   &\textcolor{blue}{4.2611} \\
    & MANIQA$\uparrow$   &\textcolor{orange}{0.5110}   &0.4292   &0.5090   &\textcolor{blue}{0.5167}   &0.5049   &\textcolor{red}{0.5307} \\

    \midrule
    \multirow{3}{*}{ Low-light + Rain \cite{cdd_dataset}}
    & MUSIQ$\uparrow$    &42.1782  &\textcolor{blue}{45.2701}  &42.8885  &42.5370  &\textcolor{orange}{43.3914}  &\textcolor{red}{51.3321} \\
    & NIQE$\downarrow$   &\textcolor{orange}{7.7417}   &\textcolor{blue}{7.4193}   &7.7843   &7.8942   &7.9027   &\textcolor{red}{6.7314} \\
    & MANIQA$\uparrow$   &0.2250   &\textcolor{orange}{0.2313}   &0.2263   &\textcolor{blue}{0.2338}   &0.2274   &\textcolor{red}{0.3095} \\

    \midrule
    \multirow{3}{*}{ Low-light + Haze \cite{cdd_dataset}}
    & MUSIQ$\uparrow$    &43.0388  &\textcolor{blue}{44.5057}  &43.4866  &43.0199  &\textcolor{orange}{44.0777}  &\textcolor{red}{52.6477} \\
    & NIQE$\downarrow$   &8.3238   &\textcolor{blue}{6.2327}   &\textcolor{orange}{7.5498}   &9.0882   &8.5612   &\textcolor{red}{6.1128} \\
    & MANIQA$\uparrow$   &0.2339   &0.2362   &\textcolor{orange}{0.2436}   &0.2290   &\textcolor{blue}{0.2442}   &\textcolor{red}{0.3342} \\

    \midrule
    \multirow{3}{*}{Low-light + Snow \cite{cdd_dataset}}
    & MUSIQ$\uparrow$    &40.2697  &\textcolor{blue}{43.7019}  &40.4844  &40.6487  &\textcolor{orange}{41.0349}  &\textcolor{red}{44.9666} \\
    & NIQE$\downarrow$   &8.3041   &\textcolor{orange}{7.7468}   &7.7828   &\textcolor{blue}{7.7434}   &8.0544   &\textcolor{red}{7.0633} \\
    & MANIQA$\uparrow$   &\textcolor{orange}{0.2176}   &\textcolor{blue}{0.2187}   &0.2099   &0.2077   &0.2149   &\textcolor{red}{0.2726} \\

    \midrule
    \multirow{3}{*}{Low-light + Haze + Snow \cite{cdd_dataset}}
    & MUSIQ$\uparrow$    &40.1995  &\textcolor{orange}{43.3599}  &41.8070  &40.3678  &\textcolor{blue}{43.3914}  &\textcolor{red}{46.4093} \\
    & NIQE$\downarrow$   &7.1736   &\textcolor{blue}{6.5793}   &\textcolor{orange}{6.5880}   &7.8782   &7.2825   &\textcolor{red}{6.1137} \\
    & MANIQA$\uparrow$   &0.2266   &\textcolor{blue}{0.2306}   &0.2238   &0.2136   &\textcolor{orange}{0.2274}   &\textcolor{red}{0.2907} \\

    \midrule
    \multirow{3}{*}{Low-light + Haze + Rain \cite{cdd_dataset}}
    & MUSIQ$\uparrow$    &41.5491  &\textcolor{blue}{45.8449}  &43.1839  &41.8434  &\textcolor{orange}{44.4468}  &\textcolor{red}{50.9273} \\
    & NIQE$\downarrow$   &7.5719   &\textcolor{orange}{7.0800}   &\textcolor{blue}{7.0241}   &7.8207   &7.6764   &\textcolor{red}{6.2831} \\
    & MANIQA$\uparrow$   &0.2483   &\textcolor{blue}{0.2573}   &\textcolor{orange}{0.2563}   &0.2403   &0.2537   &\textcolor{red}{0.3067} \\

    \midrule
    \multirow{3}{*}{Low-light + Blur \cite{lolblur}}
    & MUSIQ$\uparrow$    &44.1876  &34.9780  &\textcolor{blue}{47.8054}  &44.2333  &\textcolor{orange}{46.9724}  &\textcolor{red}{52.6950} \\
    & NIQE$\downarrow$   &5.3992   &7.5102   &5.3790   &\textcolor{blue}{5.2615}   &\textcolor{orange}{5.3741}   &\textcolor{red}{4.7135} \\
    & MANIQA$\uparrow$  &0.2316   &0.1751   &\textcolor{orange}{0.2374}   &\textcolor{blue}{0.2419}   &0.2187   &\textcolor{red}{0.2639} \\
    
  \bottomrule[1pt]
  \end{tabular}
  }
  \vspace{-2mm}
\end{table*}

\begin{figure*}[t]
\setlength{\abovecaptionskip}{1mm}
\setlength{\parskip}{0mm} 
\setlength{\baselineskip}{0mm} 
\centering
\begin{minipage}[c]{1\textwidth}
    \centering
    \vspace{-2mm}
    \includegraphics[width = 1\textwidth]{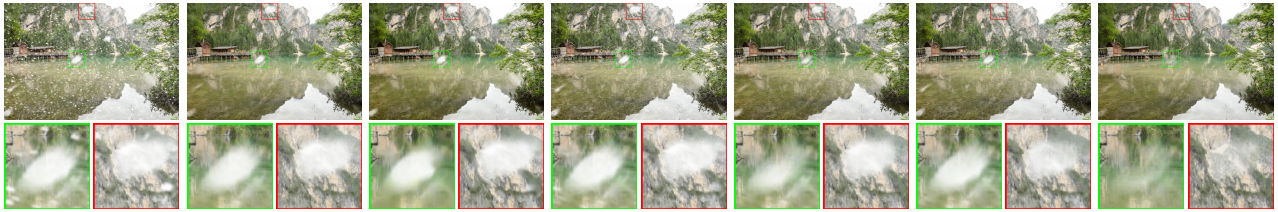}
    \includegraphics[width = 1\textwidth]{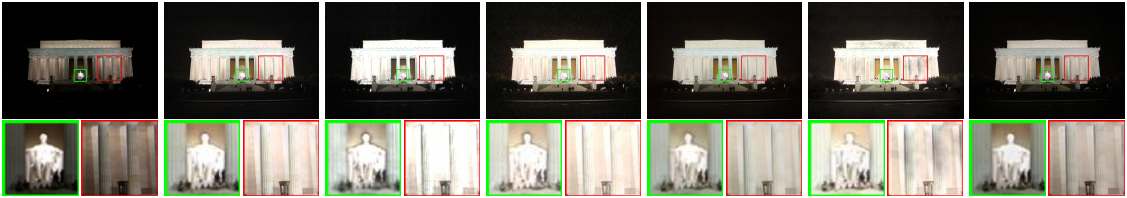}
    \includegraphics[width = 1\textwidth]{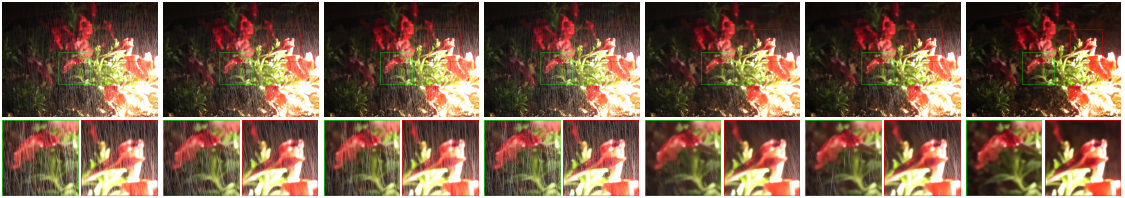}
    \begin{minipage}[c]{0.12\textwidth}
    \vspace{-1mm}
    \centering
    \scriptsize{Input}
    \end{minipage}
    \begin{minipage}[c]{0.14\textwidth}
    \vspace{-1mm}
    \centering
    \scriptsize{PromptIR}
    \end{minipage}
    \begin{minipage}[c]{0.14\textwidth}
    \vspace{-1mm}
    \centering
    \scriptsize{Histoformer}
    \end{minipage}
    \begin{minipage}[c]{0.14\textwidth}
    \vspace{-1mm}
    \centering
   \scriptsize{HOGformer}
    \end{minipage}
    \begin{minipage}[c]{0.14\textwidth}
    \vspace{-1mm}
    \centering
    \scriptsize{AdaIR}
    \end{minipage}
    \begin{minipage}[c]{0.14\textwidth}
    \vspace{-1mm}
    \centering
    \scriptsize{BioIR}
    \end{minipage}
    \begin{minipage}[c]{0.12\textwidth}
    \vspace{-1mm}
    \centering
   \scriptsize{\textbf{Ours}}
    \end{minipage}
\end{minipage}
\caption{Visual comparisons on \textit{unseen images} with a single degradation. Our method generalizes better than baselines, producing clearer outputs.}
\label{fig_setting3_unseen}
\vspace{-2mm}
\end{figure*}

\noindent \textbf{Visual Comparisons.} Fig.~\ref{fig_setting1} provides visual comparisons between our method and AWR baselines. For hazy scenes (row 1), SwinIR, PromptIR, and DCPT leave residual haze or introduce color distortion, as indicated by the red arrows. In the desnowing case (row 4), other baselines fail to eliminate snowy particles and tend to blur background structures (black arrows). In contrast, our method yields sharper contours and more faithful textures, demonstrating its robustness across diverse weather degradations. Furthermore, our method outperforms baselines under the general image restoration setting, as evidenced by clearer structures in restored blurry outputs and more vivid colors in restored low-light results in Fig.~\ref{fig_setting3}.

\noindent \textbf{Generalization on Inputs with Unseen Single Degradation.} Tab.~\ref{tab_setting3_unseen} reports quantitative comparisons between our method and SOTA baselines on unseen images with a single degradation. Models trained on Setting III are directly adopted for cross-dataset evaluation. Specifically, our method outperforms all baselines across degradations in terms of MUSIQ and MANIQA. For NIQE, our method is only slightly below PromptIR and Histoformer on certain datasets. Visual comparisons are presented in Fig.~\ref{fig_setting3_unseen}. Baseline methods struggle to produce satisfactory results, while our method is able to obtain robust degradation perceptions via our developed AWR-IQA module and produce visually appealing outputs. Our superior quantitative and qualitative performance in Tab.~\ref{tab_setting3_unseen} and Fig.~\ref{fig_setting3_unseen} highlights the importance of our devised prior-modulated and velocity-constrained rectified flow model, suggesting high applicability of our method in real-world scenarios, where captured images commonly differ from training samples.

\noindent \textbf{Generalization on Inputs with Combined Degradations.} Tab.~\ref{tab_setting3_combined} reports quantitative results on images with combined degradations, which is a more practical setting in real-world scenarios. As this setting is extremely challenging, MUSIQ and MANIQA scores suffer significant decreases compared to those reported in Tabs.~\ref{tab_setting3} and \ref{tab_setting3_unseen}, especially when low-light images are coupled with other degradations. On the other hand, it is evident that our method achieves enhanced advantages over baseline methods. Specifically, our method excels in all reported metrics, including NIQE, which differs from the trends in Tabs.~\ref{tab_setting3} and \ref{tab_setting3_unseen}. Furthermore, our average MUSIQ score across the last six tasks in Tab.~\ref{tab_setting3_combined} is significantly higher than that of the second-best method. Visual comparisons in Fig.~\ref{fig_setting3_combined} further validate the strengths of our method. Overall, these results align well with our motivation for developing the AWR-QA module, which produces soft perceptions for multiple degradations simultaneously, as well as leveraging these soft perceptions for source distribution initialization and degradation-aware velocity estimation.

\section{Ablation Studies}
\label{sec_ablation}

\begin{table}[t]
\caption{Ablations on AWR-QA, AMN, and WWA in posterior estimation stage. }
\renewcommand\arraystretch{6}
\setlength{\abovecaptionskip}{0mm}
  \centering
  \renewcommand{\arraystretch}{1.2}
  \resizebox{0.92\linewidth}{!}{
  \begin{tabular}{c|ccccc|c}
    \hline
    \textbf{Configurations} &AdaIR & Config (1) & Config (2) & Cross-attention~\cite{luo2023controlling} &AdaIR$^*$ &Our Method \\
    \hline 
    PSNR$\uparrow$ & 29.19 & 29.67  & 30.06 & 30.02 & 29.49 & 30.52\\
    SSIM$\uparrow$  &0.927 & 0.929 & 0.930 & 0.930 & 0.928 & 0.933   \\
    \hline
    \end{tabular}
  }
  \label{table_abla_stage1}
\end{table}

\begin{table}[!h]
\caption{Ablations on key components in our devised terminal-consistent residual flow.}
\renewcommand\arraystretch{1}
\setlength{\abovecaptionskip}{2mm}
  \centering
  \renewcommand{\arraystretch}{1.2}
  \resizebox{0.52\linewidth}{!}{
  \begin{tabular}{c|cccc}
    \hline
    \textbf{Configurations} &PSNR$\uparrow$ &SSIM$\uparrow$ &FID$\downarrow$  &MUSIQ$\uparrow$ \\
    \hline 
    (a) & 28.27 & 0.920  &14.55 &61.75 \\
    (b) & 28.58 & 0.922  &14.44 &61.87 \\
    (c) & 29.75 & 0.928  &11.61 &62.75 \\  
    Full Model & 30.33 & 0.932  &10.18 &63.37 \\
    \hline
    \end{tabular}
  }
  \label{table_abla_stage2}
\end{table}

Given that our method involves two stages, we conduct ablation studies in each stage to analyze the contribution of each component in our method. All ablation experiments are conducted on setting I.
\\

\noindent \textbf{Stage 1: The effectiveness of AWR-QA, AMN, and WWA for Posterior Estimation.} We firstly remove soft perceptions from AWR-QA, AMN, and WWA from our method, the architecture used for Posterior Estimation has been degraded into the AWR backbone AdaIR. We then devise several configurations: (1) only attribute perceptions and AMN are utilized; (2) only degradation perceptions and WWA are employed; (3) Both attribute and degradation perceptions are integrated using the cross-attention mechanism (DACLIP~\cite{luo2023controlling}); (4) Increase the hidden dimension of AdaIR (denoted as AdaIR$^*$) to ensure its complexity comparable to our model. Results are reported in Tab.~\ref{table_abla_stage1}. Compared to the vanilla AdaIR backbone, both the integration of attribute and weather perceptions can help achieve better posterior estimates, and more importantly, they can work complementarily to yield enhanced performance. Furthermore, compared to our method, the inferior results in configurations (3) and (4) demonstrate the advantages of our AMN and WWA modules and verify that our huge improvement is not attributed to increased model complexity but our special architectural design.

\noindent\textbf{Stage 2: Ablations on rectified-flow refinement.}
This ablation is built upon the VLM-informed posterior estimation from Stage~1.
We evaluate the roles of three components in our rectified-flow refinement: (i) perception-adaptive source perturbation, (ii) perception-aware velocity modeling, and (iii) terminal-consistent velocity parameterization.
Several configurations are compared against our full model, as summarized in Tab.~\ref{table_abla_stage2}: (a) Vanilla PMRF with a fixed perturbation scale ($\delta=0.1$).
(b) PMRF with perception-adaptive source perturbation (Sec.~\ref{sec_RF}).
(c) Built upon (b), further incorporating soft perceptions, AMN, and WWA for perception-aware velocity modeling.
Fig.~\ref{fig_rf} provides visual comparisons.
Overall, the results indicate that perception-adaptive perturbation, perception-aware velocity modeling, and the terminal-consistent velocity parameterization each contribute to the final performance.

\section{Conclusion}
\label{sec_conclusion}

In this paper, we proposed PVRF, a perception-aware rectified-flow framework for all-in-one adverse weather removal that injects VLM-extracted soft perceptions into restoration via the plug-and-play AMN and WWA modules. We further introduce a terminal-consistent residual rectified-flow design to stabilize terminal learning and improve visual realism. Extensive experiments across multiple all-in-one settings demonstrate consistent state-of-the-art performance and strong generalization to unseen and mixed degradations.

\newpage
\bibliographystyle{plain}
\bibliography{main0.bib}

\newpage
\appendix
\section{Related Work}
\label{sec_related}

\noindent \textbf{Learning-based All-in-One Adverse Weather Removal.}
Early adverse weather removal methods mainly focus on single-weather settings~\cite{mohamhaze}, where specialized models often achieve strong performance but generalize poorly to real-world scenes containing unknown or mixed degradations. To improve versatility, all-in-one restoration models have been developed in both non-blind and blind settings. Early efforts such as All-in-One~\cite{li2020all} formulate weather removal with task-aware branches, while subsequent methods explore shared and weather-specific priors for blind restoration~\cite{wswg}. More recent approaches introduce stronger feature modeling and adaptive conditioning, including transformer-based weather querying~\cite{valanarasu2022transweather}, latent-space priors~\cite{ye2023adverse}, degradation-aware prompts~\cite{PromptIR,PIP,MiOIR}, and scene- or quality-driven modulation strategies~\cite{AlwaysClearDays,cdd_dataset,AWRaCLe}. These methods substantially improve the flexibility of all-in-one restoration. However, most of them primarily rely on visual features, learned prompts, or implicit degradation descriptors, and thus may still struggle when different weather types exhibit similar perceptual characteristics or appear in coupled and varying severities. This limitation suggests the need for a more reliable perception mechanism that can explicitly characterize weather attributes and guide restoration in a degradation-aware manner.

\noindent \textbf{VLM-informed Image Restoration.}
Recent image restoration works have increasingly incorporated natural-language supervision to provide high-level semantic guidance for restoration. Existing approaches combine textual instructions with CNNs, VLMs, or generative restoration frameworks to improve controllability and user interaction~\cite{conde2025instructir,bai2023textir,brooks2023instructpix2pix}. In adverse or mixed degradation scenarios, language-aware restoration has also been explored through VLM-guided unfolding~\cite{VLU-Net}, language-derived degradation maps for expert routing~\cite{yang2024language}, and region-customized instruction-guided restoration~\cite{InstructRestore}. More recently, multimodal instruction-following systems further extend restoration toward reasoning and open-ended interaction~\cite{wu2025towards}. Despite these advances, existing VLM-guided restoration methods typically use language as a coarse semantic cue, and do not explicitly address the ambiguity among weather conditions with similar visual appearances or severity levels. As a result, their guidance may become unstable or insufficiently discriminative in all-in-one adverse weather removal, where precise perception of weather type and degradation strength is particularly important.

\noindent \textbf{Diffusion- and Flow-based Image Restoration.}
Generative restoration models based on diffusion, score matching, and flow learning have recently shown strong capability in modeling complex image distributions. Score-based and SDE-based methods~\cite{song2020score,luo2023image} formulate restoration through progressive corruption and reverse-time denoising, while DDPM-style models~\cite{ho2020denoising} provide a practical discretization of this process. To improve efficiency, probability flow ODEs~\cite{song2020score}, flow matching~\cite{lipman2022flow}, and rectified flow~\cite{liu2022flow} replace stochastic reverse processes with deterministic transport learning. For restoration, PMRF~\cite{ohayon2024posterior} further connects rectified flow with posterior-mean transport, offering an elegant framework for moving posterior estimates toward clean-image distributions. Nevertheless, existing flow-based restoration frameworks are largely designed for generic restoration settings and do not explicitly account for the perceptual ambiguity, degradation-dependent perturbation, or terminal behavior control required in all-in-one adverse weather removal. These observations motivate a restoration framework that couples perception-aware posterior estimation with a weather-adaptive transport process tailored to adverse weather scenarios.

\section{Results on Setting II}
\label{sec_setting2}
\begin{table*}[t]
  \centering
  \scriptsize

  \caption{Comparison with SOTA methods on setting II. The \textcolor{red}{Best}, \textcolor{blue}{Second-best}, and \textcolor{orange}{Third-best} methods are highlighted.}
  \label{tab_setting2}
  \centering
  \setlength\tabcolsep{1pt}
  \renewcommand{\arraystretch}{1.3}
  \resizebox{\textwidth}{!}{%
  \begin{tabular}{c|c|cc|ccccccc|ccc|c}
    \toprule[1pt]
    \multirow{2}{*}{Tasks} & \multirow{2}{*}{Metrics} & \multicolumn{2}{c|}{CNN-based} & \multicolumn{7}{c|}{Transformer-based}  & \multicolumn{3}{c|}{SDE/Diffusion-based} & \textbf{Our} \\
    & & NAFNet & DCPT-NAFNet & SwinIR & PromptIR & TransWeather & Histoformer & GridFormer & AdaIR & HOGFormer & DACLIP & GPP-LLIE & UniRestore & \textbf{PVRF} \\
    \midrule

    \multirow{8}{*}{Dehazing}
    & PSNR$\uparrow$     & \textcolor{orange}{29.4842} & 29.2242 & 26.3268 & 29.3953 & 28.8925 & 27.2265 & 27.0659 & \textcolor{blue}{29.6063} & 27.0072 & 28.4270 & 28.1868 & 24.1414 & \textcolor{red}{30.2566} \\
    & SSIM$\uparrow$     & 0.9509 & \textcolor{red}{0.9604} & 0.9389 & 0.9497 & 0.9448 & 0.9418 & \textcolor{blue}{0.9535} & 0.9512 & 0.9493 & 0.9365 & 0.9462 & 0.8346 & \textcolor{orange}{0.9526} \\
    & LPIPS$\downarrow$  & 0.0291 & \textcolor{blue}{0.0246} & 0.0444 & 0.0294 & 0.0330 & 0.0339 & \textcolor{orange}{0.0260} & 0.0296 & 0.0337 & 0.0272 & 0.0368 & 0.0952 & \textcolor{red}{0.0244} \\
    & FID$\downarrow$    & 8.1313 & 6.9793 & 13.8706 & 8.4712 & 9.5456 & 10.1239 & \textcolor{orange}{6.8404} & 8.3922 & 9.0061 & \textcolor{blue}{6.2342} & 9.9210 & 18.5698 & \textcolor{red}{5.2319} \\
    & MUSIQ$\uparrow$    & 55.9530 & 56.1014 & 56.3110 & 55.8436 & 55.9373 & 55.6795 & 56.1424 & 55.9427 & 55.8376 & \textcolor{blue}{56.8943} & 55.6146 & \textcolor{orange}{56.3163} & \textcolor{red}{56.9566} \\
    & CLIPIQA$\uparrow$  & 0.2853 & 0.2598 & 0.2939 & \textcolor{red}{0.3002} & 0.2832 & 0.2561 & 0.2630 & 0.2950 & 0.2604 & 0.2700 & \textcolor{orange}{0.2987} & 0.2659 & \textcolor{blue}{0.2994} \\
    & NIQE$\downarrow$   & 4.0587 & 4.0638 & \textcolor{red}{3.8112} & 3.9647 & 3.8340 & 4.0335 & 3.9798 & 4.0271 & 4.1015 & \textcolor{blue}{3.8247} & 3.9758 & 4.4224 & \textcolor{orange}{3.8282} \\
    & MANIQA$\uparrow$   & \textcolor{orange}{0.6416} & 0.6396 & 0.6386 & 0.6403 & 0.6356 & 0.6322 & 0.6392 & \textcolor{blue}{0.6424} & 0.6415 & 0.6338 & 0.6394 & 0.6213 & \textcolor{red}{0.6502} \\
    
    \midrule
    \multirow{8}{*}{Low-light}
    & PSNR$\uparrow$     & 18.6269 & 20.2516 & 16.1077 & 20.5389 & 20.9644 & 17.4699 & 19.9056 & \textcolor{orange}{21.8332} & 19.9568 & \textcolor{blue}{22.2302} & 20.3946 & 19.6627 & \textcolor{red}{23.6032} \\
    & SSIM$\uparrow$     & 0.8383 & 0.8183 & 0.7913 & \textcolor{orange}{0.8575} & 0.8377 & 0.7956 & 0.8112 & \textcolor{blue}{0.8669} & 0.8318 & 0.8279 & 0.8517 & 0.7387 & \textcolor{red}{0.8835} \\
    & LPIPS$\downarrow$  & 0.1153 & 0.1887 & 0.1384 & \textcolor{orange}{0.0867} & 0.1176 & 0.1995 & 0.1249 & \textcolor{blue}{0.0849} & 0.1676 & 0.1151 & 0.1157 & 0.1714 & \textcolor{red}{0.0629} \\
    & FID$\downarrow$    & 43.3469 & 68.5468 & 62.8596 & \textcolor{blue}{33.1933} & 46.6892 & 73.0903 & 49.2181 & \textcolor{orange}{34.4985} & 57.8767 & 47.1134 & 43.2211 & 71.6247 & \textcolor{red}{29.0227} \\
    & MUSIQ$\uparrow$    & \textcolor{red}{70.4247} & 55.9111 & \textcolor{orange}{69.2740} & 68.9156 & 66.8728 & 58.9973 & 65.7178 & 68.9272 & 60.1414 & 69.1371 & 68.6057 & 66.7391 & \textcolor{blue}{70.1663} \\
    & CLIPIQA$\uparrow$  & \textcolor{orange}{0.5366} & 0.4023 & 0.5277 & \textcolor{blue}{0.5571} & 0.4001 & 0.4297 & 0.4743 & 0.5162 & 0.3446 & 0.4763 & 0.4415 & 0.4795 & \textcolor{red}{0.6125} \\
    & NIQE$\downarrow$   & 4.3401 & 3.9823 & 4.3985 & 4.5524 & 3.9808 & \textcolor{red}{3.8475} & 4.4756 & 4.5544 &3.8952 & 4.9906 & \textcolor{orange}{3.8835} & 4.6244 & \textcolor{blue}{3.8652} \\
    & MANIQA$\uparrow$   & 0.6575 & 0.5897 & 0.6358 & \textcolor{orange}{0.6601} & 0.6236 & 0.5912 & 0.6210 & 0.6587 & 0.5984 & \textcolor{blue}{0.6706} & 0.6452 & 0.6328 & \textcolor{red}{0.6721} \\
    
    \midrule
    \multirow{8}{*}{Deraining}
    & PSNR$\uparrow$     & 25.4421 & 26.3477 & 26.7710 & \textcolor{orange}{27.9980} & 24.1490 & 25.5034 & 25.7722 & \textcolor{red}{28.1874} & 26.3770 & 26.1871 & 27.0394 & 21.6014 & \textcolor{blue}{28.1662} \\
    & SSIM$\uparrow$     & 0.8456 & 0.8253 & 0.8580 & \textcolor{orange}{0.8829} & 0.8101 & 0.8069 & 0.8063 & \textcolor{blue}{0.8857} & 0.8346 & 0.8360 & 0.8737 & 0.6743 & \textcolor{red}{0.8863} \\
    & LPIPS$\downarrow$  & 0.1126 & 0.1648 & 0.1064 & \textcolor{orange}{0.0755} & 0.1428 & 0.1881 & 0.1470 & \textcolor{blue}{0.0733} & 0.1560 & 0.0876 & 0.1017 & 0.2255 & \textcolor{red}{0.0628} \\
    & FID$\downarrow$    & 39.5993 & 58.2514 & 36.7157 & \textcolor{orange}{27.1002} & 51.7666 & 62.6183 & 45.6402 & \textcolor{blue}{25.8660} & 55.5853 & 32.7593 & 31.7345 & 69.8020 & \textcolor{red}{19.1047} \\
    & MUSIQ$\uparrow$    & 70.0800 & 64.5522 & 69.4326 & \textcolor{orange}{70.4612} & 68.4701 & 65.4731 & 66.8084 & \textcolor{blue}{70.6815} & 65.9262 & 69.9443 & 70.0148 & 65.8471 & \textcolor{red}{71.3017} \\
    & CLIPIQA$\uparrow$  & 0.7535 & 0.6158 & 0.7252 & \textcolor{orange}{0.7829} & 0.6509 & 0.6393 & 0.6726 & \textcolor{blue}{0.7856} & 0.6749 & 0.7466 & 0.7402 & 0.6629 & \textcolor{red}{0.7941} \\
    & NIQE$\downarrow$   & 3.6333 & 3.6589 & 3.6001 & 3.5691 & 3.7411 & \textcolor{orange}{3.2971} & 3.5996 & 3.6318 & 3.3972 & \textcolor{red}{3.1016} & 3.6051 & 3.6589 & \textcolor{blue}{3.1542} \\
    & MANIQA$\uparrow$   & 0.6707 & 0.6042 & 0.6650 & \textcolor{orange}{0.6850} & 0.6153 & 0.5982 & 0.6150 & \textcolor{blue}{0.6868} & 0.6191 & 0.6761 & 0.6767 & 0.6199 & \textcolor{red}{0.6974} \\
    
    \midrule
    \multirow{8}{*}{Desnowing}
    & PSNR$\uparrow$     & \textcolor{orange}{28.2893} & 27.2178 & 26.1904 & 27.9105 & 26.7268 & 26.5015 & 26.4140 & \textcolor{blue}{28.4000} & 26.7946 & 26.5609 & 27.3473 & 21.8470 & \textcolor{red}{28.5023} \\
    & SSIM$\uparrow$     & \textcolor{orange}{0.8966} & 0.8843 & 0.8697 & 0.8919 & 0.8743 & 0.8698 & 0.8739 & \textcolor{blue}{0.8981} & 0.8801 & 0.8661 & 0.8895 & 0.6415 & \textcolor{red}{0.9005} \\
    & LPIPS$\downarrow$  & \textcolor{orange}{0.0625} & 0.0857 & 0.0935 & 0.0665 & 0.0813 & 0.1057 & 0.0945 & \textcolor{blue}{0.0608} & 0.0897 & 0.0731 & 0.0705 & 0.1819 & \textcolor{red}{0.0520} \\
    & FID$\downarrow$    & \textcolor{orange}{21.4393} & 30.7876 & 34.3009 & 22.9940 & 30.7250 & 36.8774 & 29.5384 & \textcolor{blue}{21.1764} & 33.3173 & 25.8008 & 26.4275 & 45.5577 & \textcolor{red}{17.5538} \\
    & MUSIQ$\uparrow$    & \textcolor{orange}{69.5727} & 69.5594 & 67.9504 & 68.8747 & 69.1820 & 69.3072 & 68.2443 & 69.1902 & 69.1780 & \textcolor{blue}{70.3695} & 69.0375 & 68.8181 & \textcolor{red}{70.9611} \\
    & CLIPIQA$\uparrow$  & \textcolor{orange}{0.5652} & 0.4776 & 0.5423 & 0.5621 & 0.5379 & 0.4762 & 0.4991 & \textcolor{blue}{0.5655} & 0.4839 & 0.5082 & 0.5383 & 0.4556 & \textcolor{red}{0.5751} \\
    & NIQE$\downarrow$   & 2.9698 & 3.0606 & 2.9673 & \textcolor{orange}{2.8967} & 2.9637 & 3.2416 & 2.9067 & 2.9322 & 3.0508 & \textcolor{blue}{2.8812} & 3.0507 & 3.0606 & \textcolor{red}{2.8022} \\
    & MANIQA$\uparrow$   & \textcolor{orange}{0.6760} & 0.6583 & 0.6568 & 0.6726 & 0.6580 & 0.6568 & 0.6595 & \textcolor{blue}{0.6767} & 0.6581 & 0.6758 & 0.6658 & 0.6382 & \textcolor{red}{0.6864} \\
    \bottomrule[1pt]
  \end{tabular}
  }

\end{table*}

The results on setting II are omitted in the manuscript due to paper limit and are provided in Tab.~\ref{tab_setting2}. 

As detailed in Tab.~\ref{tab_setting2}, our PVRF outperforms most CNN, Transformer, and SDE/Diffusion-based methods across all evaluated degradation tasks. In particular, PVRF demonstrates especially strong performance in FID, important for the preservation of photo-realism. In addition, our PVRF succeeds all baselines in all metrics in desnowing scenarios, indicating its effectiveness in removing strong artifacts.

\section{Complexity Discussions}
\begin{table*}[t]
  \centering
  \scriptsize
  \caption{Comparison between our method and diffusion-based approaches in terms of metrics and efficiency. FLOPs and inference time are measured at a resolution of $256 \times 256 $ on one H100 device. \vspace{-2mm}}
  \label{tab_complexity}
  \centering
  \setlength\tabcolsep{4pt}
  \renewcommand{\arraystretch}{1.3}
  \resizebox{0.8\textwidth}{!}{%
  \begin{tabular}{c|cc|cccc}
    \toprule[1pt]
    Methods &FLOPs &Inference Time & PSNR$\uparrow$ & SSIM$\uparrow$ &MANIQA$\uparrow$ & MUSIQ$\uparrow$ \\
    \midrule
    UniRestore  \cite{chen2025unirestore}   & 1.39T &0.67s & 23.156 &0.748 &0.618 &60.462 \\
    WeatherDiff \cite{patch-diffusion}   &6.07T & 2.91s &27.953 &0.901 &0.653 &62.148 \\
    DACLIP  \cite{luo2023controlling}   &12.79T & 6.13s&28.171 &0.905 &0.657 &62.971 \\
    GPPLLIE \cite{gppllie}    &1.73T & 0.83s & 28.324&0.923 &0.655 &61.985 \\
    Ours & 9.6T+0.52T &0.35s & \textbf{30.328} & \textbf{0.932} & \textbf{0.671} & \textbf{63.375} \\
    \bottomrule[1pt]
\end{tabular}
  }
\end{table*} 

Based on rectified flow, our method belongs to the class of generative restoration approaches and is naturally suited to producing perceptually favorable results. 
To provide a more comprehensive evaluation, we further compare with several diffusion-based methods and analyze the complexity of both our method and these competitors. 
Since diffusion-based methods typically rely on iterative sampling with varying numbers of sampling steps, we report in Tab.~\ref{tab_complexity} both the total FLOPs and the inference latency, where the total FLOPs are calculated as the backbone FLOPs multiplied by the number of sampling steps.

The FLOPs reported for our method in Tab.~\ref{tab_complexity} includes frozen VLM used for AWR-QA. Note that the FLOPs and latency are not directly comparable between VLM perception and restoration networks. The LLM mainly consists of dense matrix multiplications that are highly optimized on modern GPUs and can be batched across QA prompts, whereas restoration networks process high-resolution feature maps with memory-intensive convolutions, normalization, multi-scale fusion, and iterative sampling. Thus, their latency is affected not only by FLOPs, but also by memory bandwidth, kernel-launch overhead, and solver steps. We have explicitly indicated the latency of VLM in Tab.~\ref{tab_complexity} (0.10s).

\end{document}